\documentclass{article}

\usepackage[preprint]{icml2026}

\usepackage{enumitem}

\usepackage{amsmath}
\usepackage{amssymb}
\usepackage{mathtools}
\usepackage{amsthm}

\usepackage{booktabs}
\usepackage{amsthm}
\usepackage{multirow}
\usepackage{makecell}
\usepackage{siunitx}
\usepackage{amsmath}

\makeatletter

\let\algorithmiccomment\relax
\makeatother

    \usepackage{algpseudocode}
    \usepackage{caption} 
    \usepackage{amssymb}
    \usepackage{graphicx} 
    \usepackage{dashrule}
    \usepackage{arydshln} 
    \usepackage{comment}
    \usepackage{subfig}
    \usepackage{amsfonts}
    \usepackage{float}
    \usepackage{xcolor}
    \newtheorem{proposition}{Proposition}

    \usepackage{adjustbox}
    \usepackage{makecell}
    \usepackage{hyperref}
    \usepackage{wrapfig}

\newcommand{\algvariantbox}[2]{%
  \begingroup
  \setlength{\fboxsep}{3pt}%
  \fcolorbox{#1!45}{#1!7}{%
    \begin{minipage}{\dimexpr\linewidth-2\fboxsep-2\fboxrule\relax}%
      #2
    \end{minipage}%
  }%
  \endgroup
}
\newif\ifrebuttal
\ifrebuttal
  \newcommand{\rev}[1]{\textcolor{blue}{#1}}
  \newenvironment{revblock}{\begingroup\color{blue}}{\endgroup}
\else
  \newcommand{\rev}[1]{#1}
  \newenvironment{revblock}{}{}
\fi
\algrenewcommand\algorithmiccomment[1]{\hfill\textcolor{gray}{\footnotesize // #1}}

\usepackage[textsize=tiny]{todonotes}

\icmltitlerunning{Overcoming the Communication-Performance Tradeoff in LLM Pretraining}

\begin{document}

\twocolumn[
  \icmltitle{Overcoming the Communication-Performance Tradeoff in LLM Pretraining}

  \icmlsetsymbol{equal}{*}

  \begin{icmlauthorlist}
    \icmlauthor{Amir Sarfi}{covenant}
    \icmlauthor{Benjamin Th\'erien}{udem,mila}
    \icmlauthor{Joel Lidin}{covenant}
    \icmlauthor{Eugene Belilovsky}{udem,mila,concordia}
  \end{icmlauthorlist}

  \icmlaffiliation{covenant}{Covenant AI}
  \icmlaffiliation{udem}{Universit\'e de Montr\'eal}
  \icmlaffiliation{mila}{Mila - Quebec AI Institute}
  \icmlaffiliation{concordia}{Concordia University}

  \icmlcorrespondingauthor{Amir Sarfi}{amir@tplr.ai}

  \icmlkeywords{Machine Learning, ICML}

  \vskip 0.3in
]

\printAffiliationsAndNotice{}

\begin{abstract}
Communication-efficient distributed training algorithms (e.g., DiLoCo) have received considerable interest due to their benefits for training large language models (LLMs) in bandwidth-constrained settings, such as across datacenters and over the internet. While these local-update methods achieve communication reduction through reduced synchronization, they still require communicating dense model-sized pseudo-gradients, resulting in a communication bottleneck even for cross-datacenter links. While quantization is often applied to reduce the pseudo-gradient's size, in the context of LLM pre-training, existing approaches have not been able to leverage sparsification without incurring significant performance degradation. In this work, we introduce \textit{SparseLoCo}, a communication-efficient training algorithm for LLMs that \textit{can effectively leverage $\textsc{Top-}k$ sparsification} and 2-bit quantization to reach extreme sparsity in the communicated pseudo-gradient, \textit{as high as 97--99\%}, while achieving lower final loss than dense DiLoCo. In our empirical study of language model pre-training, we demonstrate that SparseLoCo's improvements over DiLoCo in performance and pseudo-gradient compression are maintained across dense model scales (178M-2B), an MoE transformer (645M-A273M), increasing the number of workers, and increasing communication intervals.
\end{abstract}

\begin{figure}
  \centering
  \includegraphics[width=\columnwidth]{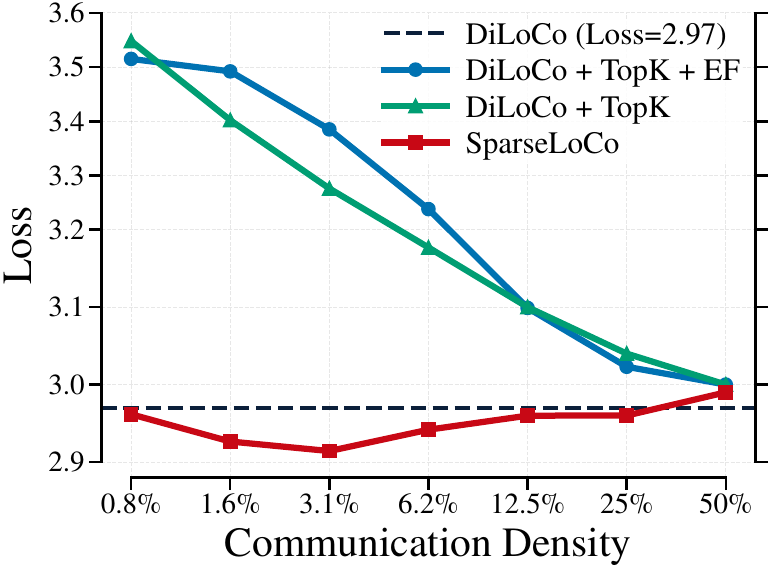}
    \caption{\textbf{Naively combining DiLoCo with \textsc{Top-}$k$ compression yields poor results, with and without error feedback. In contrast, \emph{SparseLoCo} is able to outperform DiLoCo while achieving drastic communication reduction.} We plot final validation loss as a function of communication density for a 178M model with $R{=}8$ workers and $H{=}15$ inner steps. \vspace{-3mm}
    }
    \label{fig:topk-density-sweep}
\end{figure}

\vspace{-6mm}
\section{Introduction}
Frontier language models pre-trained on internet-scale data have led to considerable breakthroughs in recent years. However, due to their growing parameter counts, effectively training these models across expensive datacenter hardware while retaining efficiency---a central goal due to the resources spent on these runs---is becoming increasingly challenging. On the other hand, with increasing availability of globally distributed computational infrastructure, the pre-training of large-scale models over the internet has recently garnered interest~\citep{intellect1}. Similar to training over the internet, pre-training across multiple datacenters requires mitigating the large communication overhead incurred by aggregating updates between workers.

\begin{figure*}[t] 
    \centering
    \subfloat[Ring Communication]{\includegraphics[width=0.4\linewidth]{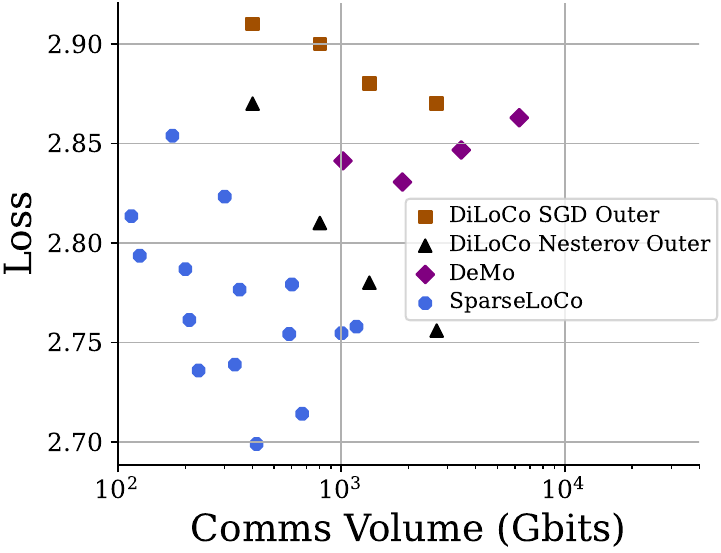}}\qquad\qquad
    \subfloat[Parameter Server]{\includegraphics[width=0.36\linewidth]{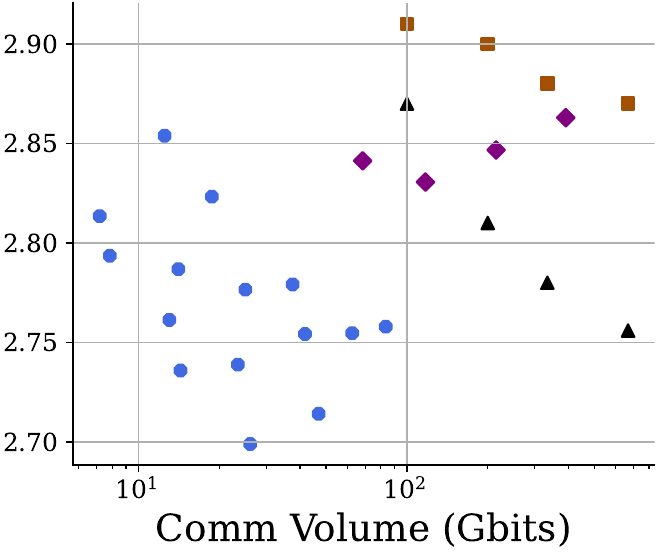}}
    \caption{\textbf{SparseLoCo lies on the Pareto frontier between loss and communication volume.} We report communication volume (outbound) for two settings (A) ring communication topology (ring all-gather for SparseLoCo and DeMo, ring all-reduce for DiLoCo) (B) Parameter server. The points consider different $H$ for DiLoCo, different densities for DeMo, and combinations of both for SparseLoCo using 512M models. We observe that, in both cases, SparseLoCo is at the Pareto frontier. }
    \label{fig:comms-volume-vs-loss}

\end{figure*}

In LLM pre-training, several approaches have been proposed to reduce data-parallel communication cost. Among them are DiLoCo~\citep{douillard2023diloco}, a variant of LocalSGD~\citep{stich2018local,reddi2020adaptive}, as well as methods compressing communicated tensors utilizing error feedback~\citep{peng2024decoupled,ahn2025dion, pmlr-v202-wang23t} to mitigate information loss. These techniques have complementary advantages of (1) reducing the communication frequency and (2) reducing the size of communicated messages. Combining the two is potentially advantageous for bandwidth-constrained settings like training over the internet or across datacenters. However, existing works focused on LLM pre-training using DiLoCo~\citep{scalingdiloco} do not take full advantage of compression schemes. 

One of the most efficient compression schemes in the literature is $\textsc{Top-}k$ sparsification; however, in this work we observe that naively combining $\textsc{Top-}k$ with DiLoCo's outer momentum leads to large performance degradation, whereas error feedback naturally acts as a local approximation of outer momentum. Building on this, we introduce \textbf{SparseLoCo}, which replaces global outer momentum with a single error feedback accumulator, thereby unifying infrequent and sparsified communication. This enables aggressive $\textsc{Top-}k$ sparsification and quantization of pseudo-gradients, while outperforming full-communication DiLoCo and frequency-compressed baselines. Our contributions can be summarized as follows. In an LLM pre-training setting:

\begin{itemize}[itemsep=2pt, parsep=0pt]
     \item We show an empirical connection between DiLoCo's outer momentum and  error feedback with sparse compression.
    \item Contrary to existing gradient compression works that \emph{only} reduce communication cost, we demonstrate that sparsifying the pseudo-gradient while treating the local momentum as an error feedback mechanism \emph{substantially improves convergence} in this setting.
    \item Leveraging these findings, we introduce SparseLoCo, a novel algorithm that blends the benefits of multi-iteration methods like DiLoCo with $\textsc{Top-}k$ sparsification, quantization, and error feedback to achieve a Pareto-optimal tradeoff between performance and communication cost.
    \item Through extensive empirical evidence, we demonstrate that SparseLoCo significantly reduces communication volume compared to existing communication-efficient LLM training methods (e.g., DiLoCo and DeMo), while simultaneously outperforming them.
\end{itemize}

\section{Related Work}

\noindent\textbf{Federated Learning}  
In the federated learning literature, communication efficiency has been a central focus from the outset, as participants often operate over highly constrained and heterogeneous networks. A canonical example is Federated Averaging (FedAvg)~\citep{konevcny2016federated,mcmahan2017communication}, which reduces communication frequency by performing multiple local updates before averaging model parameters. Other works explore compressed updates through sketching or quantization in the context of federated learning~\citep{fetchsgd,qfed}. Beyond reducing communication overhead, numerous approaches such as SCAFFOLD~\citep{karimireddy2020scaffold} and FedProx~\citep{li2020federated} address the unique challenge of data heterogeneity---where each client’s dataset may follow a different distribution---by introducing control variates or proximal terms to stabilize convergence. Related to our work, \cite{mitchell2022optimizing} consider pseudo-gradient compression in the FL setting. While our work shares federated learning’s emphasis on reducing communication overhead, it differs fundamentally in scope: we focus on large-scale pre-training of LLMs in settings with homogeneous data partitions (e.g., sharded web-scale corpora), where heterogeneity-mitigation strategies are unnecessary while achieving performance that can match standard data parallel schemes at equivalent FLOPs is paramount \cite{douillard2023diloco}.

\noindent\textbf{LocalSGD and extensions to LLM training}  
Local Stochastic Gradient Descent (LocalSGD)~\citep{stich2018local} is a widely studied approach for reducing communication in distributed training by allowing workers to perform multiple local updates before synchronizing. \cite{stich2018local} formally introduced the method and proved its convergence, while~\cite{lin2018local} highlighted that LocalSGD can lead to improved generalization compared to simply increasing the batch size. Extensions of LocalSGD include SlowMo~\citep{wang2019slowmo}, which incorporates a slow outer momentum to stabilize training in datacenter-style environments---while still using SGD as the inner optimizer---and meta-learning approaches~\citep{metalearningcomm} that adapt the aggregation function for improved performance. However, these approaches were not shown to scale well to pre-training in~\cite{ortiz2021trade}. More recently, DiLoCo~\citep{douillard2023diloco} adapted the LocalSGD framework to Large Language Model (LLM) pre-training, demonstrating that replacing the inner optimizer with AdamW and a Nesterov momentum outer optimizer can yield substantial benefits. Our work builds upon this line of research by enabling aggressive $\textsc{Top-}k$ sparsification of the communicated pseudo-gradients in a LocalSGD-style framework, something that prior methods have not achieved while maintaining or improving upon state-of-the-art LLM training performance. Other recent works focus on studying DiLoCo variants that also communicate optimizer states, allowing them to prove convergence and improve performance at the cost of additional communication~\citep{cheng2025convergence,iacob2026desloc,iacob2026mtdao,lordo}; given our focus on extreme communication efficiency, we build upon the DiLoCo line of research and choose not to communicate optimizer states. Finally, \cite{streamingdiloco} and~\cite{fournier2024wash} are orthogonal approaches that communicate a small subset of model parameters more frequently. SparseLoCo, on the other hand, maintains the benefit of infrequent communication while communicating a small-sized message and maintaining model synchronization and can be combined with approaches like \cite{streamingdiloco}. 

\noindent\textbf{Error Feedback and Compressed Updates}  
Error feedback (EF) has been extensively studied, particularly from a theoretical perspective, as a means to compensate for the information loss introduced by various gradient compression methods~\citep{seide20141bit,karimireddy2019error,stich2019error}. It has been combined with various compression techniques, including quantization, sparsification~\citep{shi2019topk}, and low-rank approximation in~\citep{powersgd,ahn2025dion}. In the single local step setting, it has been applied to LLMs in recent works~\citep{pmlr-v202-wang23t,peng2024decoupled, zhao2025separate}.
EF21-SGDM~\citep{ef21} analyzed how to combine error feedback with momentum, introducing a momentum-compatible variant that requires two accumulators and is largely focused on theoretical aspects and does not address the multi-iteration setting or the practical challenges of LLM pre-training. DeMo~\citep{peng2024decoupled} considers  a novel DCT encoding scheme alongside $\textsc{Top-}k$ compression combined with momentum SGD in the LLM setting, demonstrating it can achieve competitive performance, but without incorporating local updates or the ability to leverage adaptive optimizers. Similarly, CocktailSGD~\citep{pmlr-v202-wang23t} uses error feedback with multiple compression operators in an LLM fine-tuning setting, yet does not explore the integration of local iteration methods. Finally, QSparseLocalSGD~\citep{basu2019qsparselocalsgd} is, to our knowledge, one of the few works that combines multi-iteration methods such as LocalSGD with error feedback, but is focused on theoretical analysis with non-adaptive optimizers and without outer momentum, which are crucial to high performance in the LLM setting. In contrast, our work targets the LLM pre-training regime and develops a method to combine aggressive $\textsc{Top-}k$ compression and error feedback as an efficient approximation of the outer momentum which can improve performance while reaching extreme sparsity.

\vspace{-2mm}

\section{Methodology}

In this section, we first review DiLoCo. We then propose replacing the global outer momentum in DiLoCo with per-replica local outer momentum (LOM), where each replica maintains its own accumulator, an approach that will be used to empirically analyze the need for global momentum. Finally, we present our proposed method, \textbf{SparseLoCo}, which combines $\textsc{Top-}k$ compression with DiLoCo’s infrequent communication.

\subsection{Background and Notation}
Consider the DiLoCo/FedOpt~\citep{douillard2023diloco,reddi2020adaptive} framework, which utilizes the following basic rule on each worker or replica to produce a pseudo-gradient at each outer step, $\Delta_r^{(t)}$, as follows:
\[
\begin{aligned}
&&& \theta_r^{(t)} \leftarrow \mathrm{InnerOpt}_H\!\left(\theta^{(t-1)};\,\mathcal{D}_r\right),\quad \forall r\in[R], \\
& && \Delta_r^{(t)} \leftarrow \theta^{(t-1)}-\theta_r^{(t)} . \\
\end{aligned}
\]

Here, $H$ corresponds to the number of inner steps of the optimizer (typically AdamW), and $R$ is the number of replicas. DiLoCo, which corresponds to an instantiation of FedOpt with AdamW as the inner optimizer and outer (server) momentum using Nesterov~\citep{dozat2016incorporating}, is given as follows: 
\vspace{-9mm}

\begin{equation}\label{eq:diloco}
\begin{aligned}
& \bar\Delta^{(t)} \leftarrow \tfrac{1}{R} \sum_{r=1}^R \Delta_r^{(t)}, \\
& m^{(t)} \leftarrow \beta\, m^{(t-1)} + \bar\Delta^{(t)},\quad
      \tilde\Delta^{(t)} \leftarrow \bar\Delta^{(t)}  + \beta\, m^{(t)}, \\
& \theta^{(t)} \leftarrow \theta^{(t-1)} - \alpha\, \tilde\Delta^{(t)} .
\end{aligned}
\end{equation}

\vspace{-3mm}
where the second line corresponds to the Nesterov momentum; empirically, this has been shown to improve results~\citep{douillard2023diloco}, and omitting it reduces the method to what is often called \emph{LocalAdam} when Adam is used as the inner optimizer.

\vspace{-1mm}
\subsection{Local Outer Momentum}\label{sec:lom}
\vspace{-1.5mm}
We first propose a variant of DiLoCo that utilizes a per-replica local outer momentum instead of the unified global momentum. The goal of this algorithm is to provide insight into how well the outer momentum can be locally approximated. We denote this algorithm DiLoCo-LOM (Local Outer Momentum):
\[
\begin{aligned}
& && m_r^{(t)} \leftarrow \beta\, m_r^{(t-1)} + \Delta_r^{(t)}, \quad
      \tilde\Delta_r^{(t)} \leftarrow \Delta_r^{(t)} + \beta\, m_r^{(t)}, \\
& && \tilde{\Delta}^{(t)} \leftarrow \tfrac{1}{R} \sum_{r=1}^R \tilde\Delta_r^{(t)}, \\
& && \theta^{(t)} \leftarrow \theta^{(t-1)} - \alpha\, \tilde{\Delta}^{(t)} .
\end{aligned}
\]
Here, the outer momentum is updated locally, solely based on the local pseudo-gradient, while the final update is based on the average of the local momentum accumulators $\tilde{\Delta}^{(t)}$. Note that typical implementations of DiLoCo store the outer momentum locally on each replica, meaning that DiLoCo-LOM does not add any memory overhead compared to the global momentum variant. We show that the DiLoCo-LOM update exactly matches the DiLoCo update in Appendix~\ref{app:lom-equivalence}.

Building up to SparseLoCo, we consider an additional method, denoted DiLoCo-LOM-Sub-$k$, where the local momenta have their largest components removed at the end of each outer step:
\[
\begin{aligned}
& && m_r^{(t)} \leftarrow m_r^{(t)} - \textsc{Top-}k\!\left(m_r^{(t)}\right)
\end{aligned}
\]
This allows us to study the impact of $\textsc{Top-}k$ subtraction used in error feedback, without sparsifying the pseudo-gradient. 
\vspace{-2mm}
\subsection{SparseLoCo: Sparse Aggregation meets Local Outer Momentum}

We now introduce SparseLoCo, which blends $\textsc{Top-}k$ sparsification and error feedback in place of the local outer momentum. We consider error feedback, $e_r$, applied to the pseudo-gradients:
\begin{equation}
\label{eq:outeref}
\begin{aligned}
& e_r^{(t)} \leftarrow \beta\, e_r^{(t-1)} + \Delta_r^{(t)} \\
& \hat{\Delta}_r^{(t)} \leftarrow Q\!\left(\textsc{Top-}k\!\left(e_r^{(t)}\right)\right), \quad
  e_r^{(t)} \leftarrow e_r^{(t)} - \hat{\Delta}_r^{(t)} \\
& \Delta^{(t)} \leftarrow \tfrac{1}{R} \sum_{r=1}^R \hat\Delta_r^{(t)},  \\
& \theta^{(t)} \leftarrow \theta^{(t-1)} - \alpha\, \Delta^{(t)} .
\end{aligned}
\end{equation}
\vspace{-4mm}

Here, $Q$ is the quantization function which allows further compression of selected values. When $k$ is sufficiently small, OuterEF closely approximates the local outer momentum in LOM, since only a few components will be subtracted from $e_r$. On the other hand, unlike LOM and LOM-Sub-$k$, SparseLoCo only aggregates quantized sparse vectors, drastically reducing the message size needed for communication. The full algorithm for SparseLoCo is given in Algorithm~\ref{alg:SparseLoCo}. 

\begin{algorithm}[h]

\caption{SparseLoCo}
\label{alg:SparseLoCo}
\begin{algorithmic}[1]
  \Require  
    initial parameters $\{\theta_r^{(0)}\}$, 
    inner steps $H$, outer steps $T$, 
    outer learning rate $\alpha$, error momentum $\beta$, workers $R$, per worker training data $D_r$, and initial error buffers $\{e_r^{(0)}{=}0\}_{r=1}^R$. \vspace{2mm}
  \For{$t \gets 1$ \textbf{to} $T$}
    \ForAll{$r \in [R]$}
    \vspace{0.5mm}
      \Statex \colorbox{blue!20}{%
        \parbox{\dimexpr\linewidth-2\fboxsep\relax}{%
          \centering\textbf{Local inner loops}%
        }%
      }
      \State $\theta_{r}^{(t)} \gets \theta_r^{(t-1)}$
      \For{$h \gets 1$ \textbf{to} $H$} \Comment{Local inner loops}
        \State Sample $x\sim \mathcal{D}_r$
        \State $L\gets f(x,\theta_{r}^{(t)})$
        \State $\theta_{r}^{(t)}\gets \mathrm{AdamW}(\theta_{r}^{(t)},\nabla L)$
      \EndFor\vspace{2pt}
      \State $\Delta_r^{(t)}\gets \theta_r^{(t-1)}-\theta_{r}^{(t)}$ \Comment{Pseudo-gradient}
      \Statex \strut
      \vspace{-3mm}
      \Statex \colorbox{green!20}{%
        \parbox{\dimexpr\linewidth-2\fboxsep\relax}{%
          \centering\textbf{Compression + Error Feedback}%
        }%
      }
      \vspace{-2mm}
      \State $\hat{\Delta}_r^{(t)}\gets Q(\textsc{Top-}k(\beta\,e_r^{(t-1)}+\Delta_r^{(t)}))$
      \vspace{1mm}
\State $e_r^{(t)}\gets \beta\,e_r^{(t-1)}+\Delta_r^{(t)}-\hat{\Delta}_r^{(t)}$\label{line:ef-subk}
      \Statex \strut
      \vspace{-3mm}
      \Statex \colorbox{orange!20}{%
        \parbox{\dimexpr\linewidth-2\fboxsep\relax}{%
          \centering\textbf{Aggregate + Outer Update}%
        }%
      }
   \vspace{-2mm}
      \State $\Delta^{(t)} \gets \frac{1}{R} \sum_{s=1}^R \hat{\Delta}_s^{(t)}$\label{line:aggregation}
      \vspace{1mm}
      \State $\theta_r^{(t)}\gets \theta_r^{(t-1)} - \alpha\,\Delta^{(t)}$
            \EndFor
  \EndFor
\end{algorithmic}

\end{algorithm}

\begin{table*}[h]
\centering
\caption{SparseLoCo compared to DiLoCo and gradient compression (DeMo). We show the size of the pseudo-gradients sent, the number of synchronizations, quantization supported, and loss. SparseLoCo outperforms other communication-efficient baselines in both communication efficiency and loss. All results are reported for 512M models pre-trained on a 10B-token budget with $R{=}8$ replicas and $H{=}15$ inner steps. Here, \emph{Density} denotes the communication density as the percentage of coordinates in the pseudo-gradient vectors that are non-zero (and therefore transmitted) at each synchronization step.\vspace{-1mm}}\label{tab:main}
\begin{tabular}{lccccc}
\toprule
\textbf{Method} & \textbf{Density} & \textbf{Loss} & \textbf{Pseudo-Grad Size} & \textbf{\# of Syncs} & \textbf{Quantization} \\
\midrule
AdamW DDP  & 100\% & 2.69 & 1.02 GB & 2445 & 16-bit \\
\noalign{\vskip 3pt}\cdashline{2-6}\noalign{\vskip 3pt}
DiLoCo (H=15) & 100\% & 2.76 & 512.40 MB & 163 & 8-bit \\
DeMo & 0.78\% & 2.83 & 10.01 MB & 2445 & 8-bit \\
DeMo  & 3.12\% & 2.86  & 40.03 MB & 2445 & 8-bit \\
SparseLoCo (H=15) & 0.78\% & 2.79 & 5.45 MB & 163 & 2-bit \\
SparseLoCo (H=15) & 3.12\% & \textbf{2.70} & 17.21 MB & 163 & 2-bit \\
\bottomrule
\end{tabular}
\end{table*}

SparseLoCo uses a \rev{chunk-wise} variant of the $\textsc{Top-}k$ operation inspired by~\cite{9546514,peng2024decoupled}. \rev{To do so, we partition each 2D parameter tensor (e.g., attention and MLP weight matrices) into non-overlapping $64\times 64$ blocks and each 1D tensor (e.g., layer-norm parameters) into contiguous chunks of size $4096$, and then apply $\textsc{Top-}k$ independently within each chunk.} This has three benefits compared to applying it at the full-tensor or global level: (a) the cost of naively storing indices for transmission is significantly reduced as each chunk's index space is bounded (further discussed in Appendix~\ref{app:compression_size}). (b) $\textsc{Top-}k$ applied to entire models or individual tensors can overemphasize correlated variables; thus, chunking can have benefits on performance as further discussed in Appendix~\ref{sec:chunking}. (c) Finally, chunking simplifies integration with tensor parallelism (TP) and Fully Sharded Data Parallel (FSDP) training, where parameters are partitioned across devices, by avoiding inefficient global $\textsc{Top-}k$ over entire parameters.

\section{Experiments}
Our experiments use 178M-, 512M-, and 2B-parameter LLaMA-style decoder-only transformers on DCLM~\citep{dclm} using the LLaMA-2 tokenizer~\citep{touvron2023llama}. Following~\cite{hoffmann2022training}, we allocate a token budget equal to \(20\times\) the model size. Our experimental protocol follows~\citep{scalingdiloco}. Unless otherwise stated, our main results are reported on the 512M-parameter model with $R{=}8$ workers, per-worker batch size \(B{=}256\), and sequence length \(L{=}2048\), yielding a global batch of \(B \times L \times R \approx 4.19\)M tokens per step with $H{=}15$ communication interval. For SparseLoCo, we employ 2-bit quantization with a chunk size of \(4096\) (non-overlapping square $64\times64$ grids for 2D parameters) and error-feedback coefficient $\beta=0.95$. After chunk-wise $\textsc{Top-}k$, we quantize all selected values from the same tensor together. We follow the statistical quantization scheme used in Intellect-1~\citep{intellect1}: values are centered by their mean, the range $[-6\,\sigma,6\,\sigma]$ is divided into uniform bins (four bins for 2-bit quantization), and each bin stores the centroid of its assigned values as a lookup table. We dequantize the transmitted values before aggregation and the outer update. We apply a short error feedback (OuterEF) freeze, where the error feedback $e_r$ is not utilized for the first $5\%$ of the outer steps to improve training stability and performance (ablated in Table~\ref{tab:ef-freeze} in Appendix~\ref{app:ef-freeze}). We also ablate EF decay $\beta$ in Appendix~\ref{app:ef-decay}. We further study scaling across model sizes (178M; Appendix~\ref{app:sacling-R}, 2B; Table~\ref{tab:2B}), and number of workers \(R\in\{16,32\}\) in Table~\ref{tab:workers_H50}, and Appendix Tables~\ref{tab:178M_scaling_r}, and~\ref{tab:512M_scaling_r}. We report the hyperparameter sweep ranges, architectural details, and selected configurations in Appendix~\ref{app:sweeps}. As baselines, we include DiLoCo and DeMo---two strong communication-efficient methods (one using local iterations and the other using compression with error feedback) for LLMs---as well as a DDP AdamW baseline.

\vspace{-3mm}

\subsection{\textsc{Top-}$k$ compression and EF with Local Optimizers and LLMs}\label{sec:topk-ef-localadam}
\begin{figure}[h]
  \centering
  \vspace{-1mm}
  \includegraphics[width=.999\columnwidth]{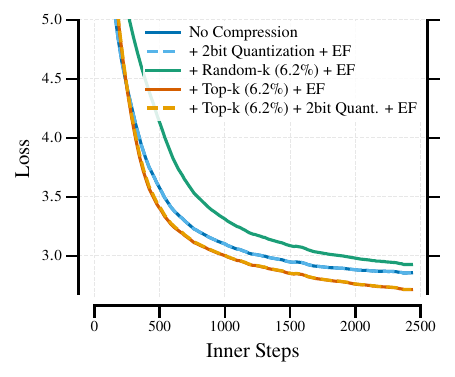}\vspace{-4mm}
  \caption{\textbf{\textsc{Top-}$k$ sparsification with error feedback improves LocalAdam in LLM pre-training.} We plot validation loss for LocalAdam (DiLoCo with SGD outer optimizer) under different pseudo-gradient compressors combined with error feedback. Quantization does not impact performance (curves closely overlap full-precision counterparts), Random-$k$ slightly degrades, while \textsc{Top-}$k$ substantially improves both convergence speed and final loss. All results report 512M LLaMa-2 models pre-trained on a 10B-token budget with $R{=}8$ replicas and $H{=}15$ inner steps.}
  \label{fig:localadam-loss-convergence}
\end{figure}
We start by demonstrating the unique properties of \textsc{Top-}$k$ compressors combined with error feedback in the setting of LLMs and local optimizers.  We consider the DiLoCo with SGD outer optimizer or, as it is referred to in the literature, LocalAdam. Similar to \cite{basu2019qsparselocalsgd}, which uses CNNs and SGD local optimizers we apply a number of compressors with full-memory error feedback ($\beta{=}1$). This corresponds to using different compressors in \eqref{eq:outeref}.  The results are shown in Figure~\ref{fig:localadam-loss-convergence}. As confirmed in prior literature, we observe that combining error feedback and compression can lead to reduced communication cost while maintaining the performance or with modest convergence degradation.  Specifically, with aggressive quantization (2-bits), performance does not degrade, while random-k sparsification with $6.25\%$ density leads a slight degradation. On the other hand, we observe that a unique phenomenon arises when \textsc{Top-}$k$ sparsification is combined with error feedback: both final performance and convergence speed improve substantially. We emphasize that this is a unique property of the combination of \textsc{Top-}$k$ and EF that to the best of our knowledge has not been observed in the literature on compression and error feedback.

\vspace{-1.5mm}
We hypothesize that this improvement is due to the error feedback buffer providing benefits similar to the outer momentum in DiLoCo. Motivated by this result, we further investigate this connection, and then evaluate SparseLoCo's performance against strong baselines across multiple settings and provide a thorough ablation study.

\vspace{-3mm}
\subsection{Building intuition with Local Outer Momentum}
\vspace{-1mm}

\begin{table}[h]
  \centering
  \caption{\textbf{DiLoCo’s global outer momentum is well approximated by sparse local outer momentum.}\vspace{-1mm}}
  \label{tab:diloco_lom_results}
  \begin{tabular*}{0.8\linewidth}{@{\extracolsep{\fill}}l c}
    \toprule
    Method & Loss \\
    \midrule
    DiLoCo & 2.760 \\
    DiLoCo w.o.\ outer momentum & 2.868 \\
    DiLoCo-LOM & 2.759 \\
    DiLoCo-LOM-Sub-$k$ (25\%) & 2.761 \\
    \bottomrule
    \vspace{-5mm}
  \end{tabular*}
\end{table}

We use the DiLoCo-LOM and DiLoCo-LOM-Sub-$k$ algorithms to empirically link the standard outer momentum in DiLoCo to the error feedback mechanism in SparseLoCo. In DiLoCo-LOM, each replica maintains a local outer momentum accumulator that is averaged only at synchronization. In DiLoCo-LOM-Sub-$k$, we subtract the largest entries of the local momentum after each synchronization. Here, the local momentum is identical to the error feedback in $\textsc{Top-}k$ compression as it is both maintained locally, and the $\textsc{Top-}k$ largest magnitude entries are constantly removed, while keeping the communicated pseudo-gradients dense. 
As shown in Table~\ref{tab:diloco_lom_results}, DiLoCo-LOM matches DiLoCo’s loss---consistent with Proposition~\ref{prop:lom_equals_global} (Appendix~\ref{app:lom-equivalence})---and pruning $25\%$ of the largest accumulator entries has a negligible impact, whereas removing outer momentum entirely degrades performance. This experiment shows momentum benefits are maintained even with aggressive pruning of the largest values in the outer momentum or error feedback.

\vspace{-2mm}
To quantify how closely local outer momentum tracks the target global outer momentum in DiLoCo, we maintain a reference global accumulator and, over the first $20$ outer steps, compute the cosine similarity between this reference and each replica’s local accumulator at corresponding steps. The average similarity between individual local accumulators and the global reference accumulator is $\ge 0.75$ for DiLoCo-LOM-Sub-$k$ (25\%), indicating that removing the top components from local accumulators each outer step still allows them to remain a strong directional proxy for the global momentum and supporting our interpretation of SparseLoCo’s error feedback state as a local approximation of DiLoCo's outer momentum.

\begin{figure*}[t]
    \centering
    \includegraphics[width=0.45\linewidth]{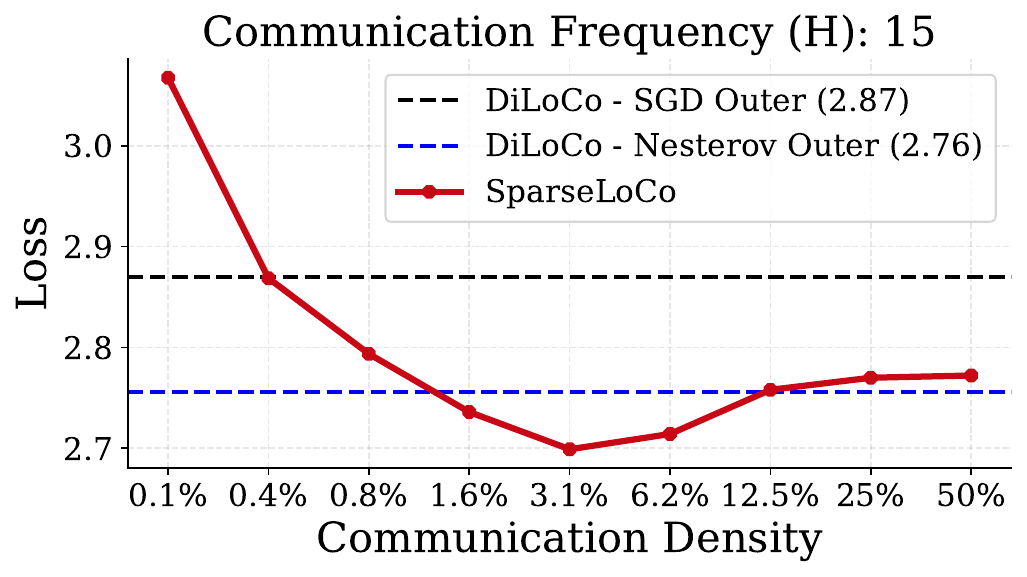}
    \includegraphics[width=0.45\linewidth]{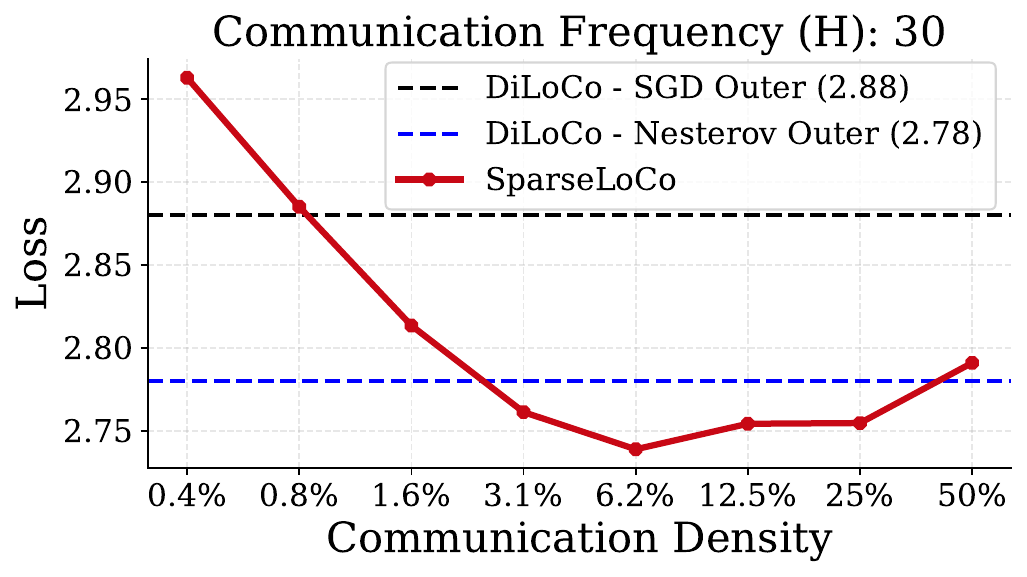}
    \includegraphics[width=0.45\linewidth]{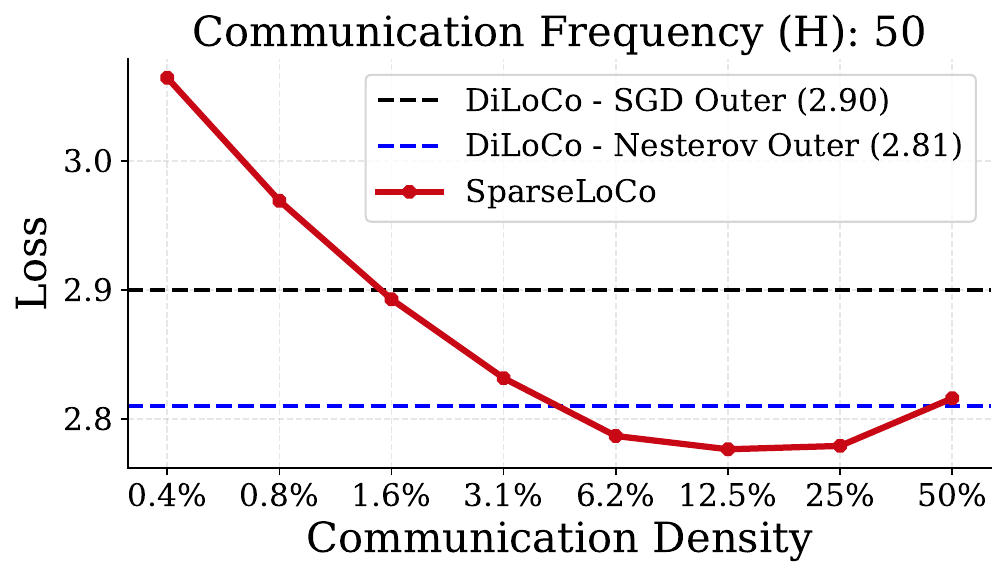}
    \includegraphics[width=0.45\linewidth]{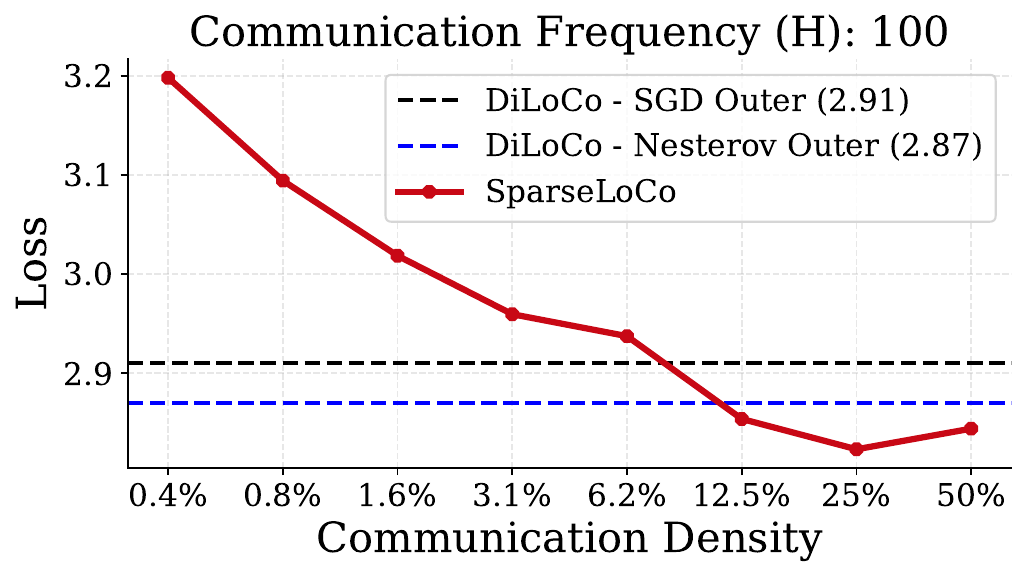}
    \caption{\textbf{SparseLoCo outperforms DiLoCo for $H\in\{15,30,50,100\}$ communication intervals.} We evaluate SparseLoCo, DiLoCo, and DiLoCo with SGD outer optimizer across different communication intervals, and different sparsity levels for SparseLoCo. We report the best performance in each case. Crucially, SparseLoCo can outperform DiLoCo while communicating significantly less. We also observe that the optimal density grows with higher communication intervals. All experiments were conducted using 512M models.
    }\label{fig:comm-freq}
\end{figure*}

 \vspace{-3mm}
\subsection{Naive \textsc{Top-}$k$ for DiLoCo}\label{sec:naive-topk-diloco}
\vspace{-1mm}
We now present two straightforward ways to leverage \textsc{Top-}$k$ sparsification into DiLoCo and explain why they do \textbf{not} work.
The first (DiLoCo + \textsc{Top-}$k$) simply applies the \textsc{Top-}$k$ operation before communication.
The second (DiLoCo + \textsc{Top-}$k$ + EF) adds error feedback to track any non-communicated information.
These modifications correspond to applying \textsc{Top-}$k$ and an optional error feedback to the local pseudo-gradients immediately before Eq.~\ref{eq:diloco}. %

\vspace{-1mm}
In Figure~\ref{fig:topk-density-sweep}, we plot final loss as a function of sparsity for SparseLoCo and these naive \textsc{Top-}$k$ DiLoCo variants. We find that SparseLoCo using only a local EF can outperform all methods at low sparsity, including dense DiLoCo. In contrast, the performance of DiLoCo + \textsc{Top-}$k$ and DiLoCo + \textsc{Top-}$k$ + EF substantially degrades at high sparsity.
We hypothesize that this is caused by the outer momentum preserving directions corresponding to the large magnitude components, which then dominate subsequent updates.

\vspace{-3mm}
\subsection{SparseLoCo Evaluations}

\begin{table}[h]
\centering
\vspace{-2mm}
\caption{Benchmark (0-shot) accuracy (higher is better), \textbf{Best} is bold. 
We evaluate the same 512M pretrained models as in Table~\ref{tab:main}, using best performing models for SparseLoCo and DeMo. We observe that SparseLoCo outperforms the DeMo and DiLoCo baselines across all benchmarks.}
\label{tab:bench-acc}
\begin{tabular}{lccc}
\toprule
\textbf{Method} & \textbf{ARC-Easy} & \textbf{HellaSwag} & \textbf{PIQA} \\
\midrule
AdamW DDP         & 44.99\% & 36.08\% & 65.34\% \\
\noalign{\vskip 3pt}\cdashline{2-4}\noalign{\vskip 3pt}
DiLoCo     & 44.28\% & 34.50\% & 64.96\% \\
DeMo              & 41.92\% & 32.37\% & 64.09\% \\
SparseLoCo  & \textbf{45.24\%} & \textbf{36.49\%} & \textbf{65.23\%} \\
\bottomrule
\vspace{-6mm}
\end{tabular}
\end{table}
\vspace{-2pt}
Table~\ref{tab:main} compares SparseLoCo at $H{=}15$ against existing state-of-the-art methods for LLMs. We utilize 2-bit quantization for SparseLoCo with no observed loss degradation, while using the prescribed quantization settings for baselines~\citep{douillard2023diloco,peng2024decoupled}. We observe that SparseLoCo obtains lower final loss than DiLoCo and DeMo baselines, while enjoying the simultaneous communication benefits of aggressively sparsified pseudo-gradients and reduced synchronization frequency. As SparseLoCo inherently utilizes error feedback, SparseLoCo further reduces communication size by quantizing the sparsified values. We further compare the performance on simple downstream tasks relevant at this model scale in Table~\ref{tab:bench-acc}, demonstrating that the performance improvements are consistent.

\vspace{-3mm}
\paragraph{SparseLoCo performance at different sparsity levels and communication intervals.} In Figure~\ref{fig:comm-freq}, we further demonstrate the performance improvements of SparseLoCo \rev{across $\textsc{Top-}k$ densities and} increasing $H\in\{15,30,50,100\}$ values compared to well-tuned DiLoCo and DiLoCo without outer momentum baselines. We observe a trend that aligns with the hypothesis that SparseLoCo's OuterEF can provide similar benefits to DiLoCo's outer momentum. In particular, we first observe that not using outer momentum in DiLoCo leads to significant performance degradation and that this setting corresponds exactly to the fully dense case for $\textsc{Top-}k$ (e.g., $k{=}100\%$). With SparseLoCo, we observe that (i) extreme sparsity levels (when nearly nothing is sent, e.g., $0.05\%$) degrade performance. (ii) With increasing density (while remaining sparse), performance improves and eventually exceeds DiLoCo for all values of $H$. \rev{At these density levels, the EF buffer remains relatively dense and accumulates residual gradients, resembling a sparsified outer momentum in accumulating gradients} (iii) Finally, as $k$ approaches dense communication, the EF buffer becomes more sparse (due to line~\ref{line:ef-subk} in Algorithm~\ref{alg:SparseLoCo}), trending towards the performance of DiLoCo with no outer momentum and thus degrading performance. The same three regimes appear across all communication intervals $H$. Furthermore, in Figure~\ref{fig:demo-density} in Appendix~\ref{app:demo-density}, we observe the same phenomenon with DeMo~\citep{peng2024decoupled} as well, while the overall performance is inferior to SparseLoCo.

\begin{table*}[t]
\small
\centering
\caption{Evaluation loss and benchmark (0-shot) accuracy of 2B-parameter LLMs with $R{=}16$ contributing peers. Best in \textbf{bold}.}
\label{tab:2B}
\begin{tabular}{lcccccc}
\toprule
\textbf{Method} & \textbf{Loss} & \textbf{ARC-Easy} & \textbf{ARC-Challenge} & \textbf{HellaSwag} & \textbf{PIQA} & \textbf{WinoGrande} \\
\midrule
AdamW DDP              & 2.34 & 58.42\% & 32.00\% & 56.87\% & 72.25\% & 56.59\% \\
\noalign{\vskip 3pt}\cdashline{2-7}\noalign{\vskip 3pt}
DiLoCo    & 2.37          & \textbf{60.48}\%             & 30.55\%             & 54.95\%          & \textbf{72.85}\%             & 55.56\% \\
SparseLoCo                 & \textbf{2.36} & 59.05\%    & \textbf{32.17\%}    & \textbf{55.49\%} & \textbf{72.85\%}  & \textbf{58.56\%} \\
\bottomrule
\end{tabular}
\end{table*}

\vspace{-1mm}
\textbf{SparseLoCo can outperform DiLoCo.} Across all settings of the inner steps $H$, we observe regimes where SparseLoCo outperforms DiLoCo. A plausible explanation is that sparse aggregation at a well-chosen $k$ emphasizes high-saliency components and reduces interference among updates, echoing intuitions from recent model-merging work in multi-task fine-tuning~\citep{yadav2023tiesmerging,davari2024model}.

\vspace{-4mm}
\paragraph{Higher sparsity is needed with fewer inner steps.} 
We observe through Figure~\ref{fig:comm-freq} a systematic pattern that the optimal value of SparseLoCo is reached at a higher sparsity level with fewer inner steps. This is consistent with the fact that higher inner steps communicate information from a larger total number of samples. Indeed, we would expect that a trajectory with more steps would have a larger support.

\vspace{-4mm}
\paragraph{SparseLoCo is at the Pareto frontier in communication volume.} In Figure~\ref{fig:comms-volume-vs-loss} we compare the communication volume of SparseLoCo to DiLoCo, DiLoCo w.o. outer momentum, and DeMo. The exact communication setting and the underlying implementation of the aggregation can have a significant impact on the communication volume. We consider two common setups from the literature---methods utilize either ring all-reduce or ring all-gather (Fig. A), or a parameter server (Fig. B). We observe that in both cases, SparseLoCo lies on the Pareto frontier while other methods have a strictly worse tradeoff. We note that the results in Fig. A assume aggregation using a naive all-gather operation for implementation, while there is further potential to exploit the structure of the problem, for example, by summing overlapping indices along steps in the all-gather ring or utilizing specially designed all-reduce~\citep{li2022near}. In Section~\ref{app:deploy} of the Appendix, we also discuss the communication measured during a live deployment of collaborative learning over the internet using SparseLoCo.

\vspace{-4mm}
\paragraph{SparseLoCo can be used with Ring All-Reduce as a drop-in for DiLoCo.} Although our analysis and motivation in the work focus on aggressively compressing the per-iteration message size, we note that in communication settings where efficient all-reduce is already available and preferred, SparseLoCo still provides significant benefit over DiLoCo while incurring no additional memory or compute overhead. Concretely, the aggregation step in Algorithm~\ref{alg:SparseLoCo} \rev{Line~\ref{line:aggregation}} can be performed directly by an all-reduce over a sparse vector. This has two significant benefits over DiLoCo with all-reduce (AR): (1) As observed in Table~\ref{tab:ablate-overtraining} and Figure~\ref{fig:comm-freq}, the performance is improved when $k$ is optimally selected and (2) the outer error feedback naturally supports more aggressive quantization than the naive DiLoCo, allowing for 2-bit quantization to be used \textit{without an additional accumulator}, unlike~\cite{muloco}.

\paragraph{SparseLoCo scales across model sizes, communication intervals, and number of replicas.}
In Table~\ref{tab:workers_H50}, we evaluate scaling of DiLoCo and SparseLoCo with number of workers $R\in\{8,16,32\}$ using a 512M-parameter model scale and communication interval $H{=}50$. Scaling beyond 8 replicas without significant degradation is a known challenge \cite{scalingdiloco}. We observe that SparseLoCo consistently outperforms DiLoCo with higher number of workers across all settings and across a number of densities, showing that it can help address the challenge of scaling the number of replicas. We also observe that with higher number of workers a lower density can sometimes be supported. In the Appendix \ref{tab:178M_scaling_r} we also study the impact of the number of replicas at 178M model size. We also evaluate SparseLoCo in the highest communication intervals ($H{=}250$), for this we follow~\cite{scalingdiloco} using an overtraining regime with a doubled token budget, where we see again SparseLoCo is able to achieve competitive performance with DiLoCo while improving communication. Finally, we run a larger scale model of size 2B-parameter model scale with $R{=}16$ workers and communication interval $H{=}50$, where SparseLoCo with $6.25\%$ density outperforms DiLoCo (Table~\ref{tab:2B}). We can see that the benefits of SparseLoCo are maintained at this scale.

\begin{table}[t]
\caption{The final evaluation loss of scaling number of replicas $R\in\{8,16,32\}$ for a 512M-parameter model with communication interval $H{=}50$ under different communication densities. SparseLoCo consistently outperforms DiLoCo as R increases. Best result is presented in \textbf{bold}.}
\label{tab:workers_H50}
\centering
\begin{tabular}{lcccc}
\toprule
Method & Density & \multicolumn{3}{c}{Loss} \\
\cmidrule(lr){3-5}
 &  & $R{=}8$ & $R{=}16$ & $R{=}32$ \\
\midrule
AdamW & 100.00\% & 2.69 & 2.69 & 2.69 \\
DiLoCo & 100.00\% & 2.81 & 2.87 & 2.93 \\
\midrule
\multirow{6}{*}{SparseLoCo}
 & 0.78\%  & 2.97 & 3.00 & 3.09 \\
 & 1.56\%  & 2.89 & 2.92 & 3.00 \\
 & 3.12\%  & 2.83 & 2.86 & 2.92 \\
 & 6.25\%  & 2.79 & 2.82 & \textbf{2.88} \\
 & 12.50\% & \textbf{2.78} & \textbf{2.80} & 2.91 \\
 & 25.00\% & \textbf{2.78} & 2.84 & 3.02 \\
\bottomrule
\vspace{-4mm}
\end{tabular}
\end{table}

\paragraph{Mixture-of-Experts models.}
SparseLoCo can also be applied to MoE transformers without any algorithmic changes. On a Qwen2-MoE-style 645M-A273M model, we observe that SparseLoCo at $3.12\%$ density (loss=$2.741$) closes the gap to AdamW DDP training (loss=$2.739$) while substantially outperforming DiLoCo (loss=$2.803$). See Appendix~\ref{app:moe} for additional details and results.

\begin{table}[H]
\caption{\textbf{MoE validation loss comparison.}
We report final validation loss for the 645M-A273M Qwen2-MoE-style model. AdamW and DiLoCo use the best settings from their MoE-specific sweeps, while SparseLoCo reuses the dense-model hyperparameters from Appendix~\ref{app:sweeps}. SparseLoCo nearly matches AdamW at $3.12\%$ density and improves substantially over DiLoCo. Best result is \textbf{bolded}.}
\label{tab:moe_results}
\centering
\begin{tabular}{lcc}
\toprule
Method & Density & Loss \\
\midrule
AdamW      & 100\%  & \textbf{2.739} \\
SparseLoCo & 3.12\% & 2.741 \\
SparseLoCo & 1.56\% & 2.776 \\
DiLoCo     & 100\%  & 2.803 \\
\bottomrule
\end{tabular}
\end{table}

\paragraph{Wall-clock time and compute utilization.} Figures~\ref{fig:wallclock} and~\ref{fig:compute-util} in Appendices~\ref{app:wallclock} and~\ref{app:compute-util} provide training wall-clock time and compute utilization results, showing strong communication efficiency of SparseLoCo.

\paragraph{Large scale deployment.} SparseLoCo was used in a real-world, collaborative, globally distributed training run of a 72B-parameter LLM in a low-bandwidth environment, representing the largest deployment of this kind reported to date~\cite{intellect1}.  The downstream performance of the resulting model is competitive with similar models trained and open-sourced in high-bandwidth cluster settings. Owing to the substantial cost and logistical complexity of training at this scale, exact comparisons to centralized optimizers with identical data, token budgets are not practically feasible.  Additional deployment details and empirical measurements are provided in Appendix~\ref{app:deploy}.

\subsection{Ablations} 
We now highlight key design choices of SparseLoCo through a series of ablations.

\begin{table}[t]
\noindent
\centering
\caption{\textbf{Ablation Studies} (\textit{Top}): SparseLoCo with Random-$k$ vs. \textsc{Top-}k sparsification; \textsc{Top-}k significantly outperforms Random-$k$ across all densities. (\textit{Bottom}): Effect of quantization on loss; 2-bit shows almost no degradation vs. full precision.}
\label{tab:ablation}
\renewcommand{\arraystretch}{1.1}

\begin{adjustbox}{center,max width=\columnwidth}
\begin{tabular}{ccc}
  \toprule
  \textbf{Density} & \textbf{\shortstack{Random-$k$ Loss}} & \textbf{\shortstack{\textsc{Top-}k Loss}} \\
  \midrule
  1.56\% & 3.05 & 2.74 \\
  3.12\% & 2.98 & 2.70 \\
  6.25\% & 2.93 & 2.71 \\
  \bottomrule
\end{tabular}
\end{adjustbox}

\vspace{4pt}

\begin{adjustbox}{center,max width=\columnwidth}
\begin{tabular}{cccccc}
  \toprule
  \textbf{\shortstack{Quant.\\ bits}} & \textbf{1} & \textbf{2} & \textbf{3} & \textbf{4} & \textbf{32} \\
  \midrule
  \textbf{Loss} & 4.79 & 2.70 & 2.70 & 2.70 & 2.70 \\
  \bottomrule
  \vspace{-4mm}
\end{tabular}
\end{adjustbox}

\end{table}

\paragraph{Random-$k$.} We ablate the choice of $\textsc{Top-}k$ compared to the alternative Random-$k$~\citep{shi2019topk,pmlr-v202-wang23t} in Table~\ref{tab:ablation}. We observe that performance is significantly degraded when using random-$k$ for the same number of indices selected, emphasizing the importance of this design choice. 

\paragraph{Quantization.} As discussed, SparseLoCo benefits from stronger quantization than non-EF methods and supports up to 2-bit quantization and was generally observed to give results very close to full precision. In Table \ref{tab:ablation}, we show the performance at different quantization values, showing that 2-bit quantization can be achieved at almost no performance cost.

\section{Conclusion}
 We have proposed SparseLoCo, an algorithm that can blend multi-iteration LLM pre-training methods with $\textsc{Top-}k$ sparsification and quantization, enabling aggressive compression of DiLoCo's pseudo-gradients. Our work establishes that the outer momentum in DiLoCo can be replaced by local momentum accumulators without losing performance. Connecting local momentum with error feedback, we leverage this insight to develop \emph{SparseLoCo}.
Our extensive experiments confirm that SparseLoCo significantly reduces communication while outperforming strong baselines such as DiLoCo and DeMo, placing it on the Pareto frontier of loss versus communication volume. Additionally, our experiments reveal that sparse aggregation may actually be useful for improving performance, opening the possibility of studying more sophisticated aggregation methods in the pre-training setting.

\bibliography{ref}
\bibliographystyle{icml2026}

\clearpage

\appendix
\onecolumn

\section{Real-World Deployment for Collaborative Permissionless Distributed Training Over the Internet}\label{app:deploy}
SparseLoCo has been deployed in a real-world setting and is being used to collaboratively train models up to 8B and 72B with permissionless global participants using an incentive scheme~\cite{lidin2025incentivizing} that rewards participants purely based on analysis of their compressed pseudo-gradients.
This was done on top of an existing blockchain. The addition and coordination of peers and their rewards were handled through the blockchain. Communication of pseudo-gradients was routed through globally distributed, S3-compliant object storage---specifically Cloudflare R2---which enabled rapid dissemination of model updates worldwide. This setup allowed updates to be time-stamped and verified as part of the reward mechanism~\cite{lidin2025incentivizing}. Each peer maintained their own storage bucket, posting read credentials to a blockchain so that both other peers and the reward mechanism could access their compressed pseudo-gradients.

SparseLoCo is particularly advantageous in this communication setup, as cloud providers have high bandwidth for peer downloads and are able to rapidly distribute and mirror files across the globe. On upload, each peer sends only its pseudo-gradient through the cloud provider. Therefore, their outbound communication (and required upload bandwidth) is kept low. They then download the pseudo-gradients from the cloud provider, which is able to easily handle the high bandwidth constraints.  An example of a practical communication time measured with an 8B model is on average 12 seconds, including sending their compressed pseudo-gradients and downloading other workers' messages with the test node never exceeding 500 Mb/s. Compared to processing with $8\times$ H200, which takes around 4.5 minutes, leading to minimal wall-clock time degradation despite traffic over the internet.  For reference, \cite{intellect1}, which trained a similarly sized model (10B) with 8-bit DiLoCo, reports a globally distributed all-reduce synchronization time of 8.3 minutes on average for a peak of $R{=}14$ nodes participating and processing time of 38 minutes.  
We also performed test measurements of communication time for a 70B model with the same setup as above ($R{=}20$ peers), measuring a total communication time of 70 seconds on average, with the test node never reaching more than 500 Mb/s downlink and 110 Mb/s uplink.  

\textbf{72B LLM training}
We have deployed SparseLoCo on the live system discussed above to train a 72B model, the largest collaborative foundational model training run ever considered (roughly $7\times$ the scale of the prior largest decentralized training run \cite{intellect1} with lower communication overhead as discussed above). The deployment uses a maximum of $R{=}20$ replicas, each associated with a peer (with variation in the number of participants during training as peers join and leave), $H{=}30$ inner steps, and a global batch size of approximately 8M tokens per inner step. We utilized a peak learning rate of 1.2e-4 and the DCLM dataset. Although the system design allows peers to use any target hardware that can achieve reasonable throughput, the suggested hardware requirements targeted $8\times$ B200 GPUs per replica. Table~\ref{tab:internet-vs-centralized} compares our Internet-trained 72B model against both decentralized (Intellect-1~\cite{intellect1}) and centralized (LLaMA-2~\cite{touvron2023llama} and LLM360 K2~\cite{liu2025llm360}) large-scale baselines across standard zero-shot benchmarks. We observe that the SparseLoCo-trained model is able to achieve competitive performance particularly for the token budget.

\begin{table}[h]
\centering
\caption{\textbf{Benchmark comparison across Internet-trained and centralized-cluster-trained LLMs.}
We report zero-shot accuracy on ARC-C, ARC-E, HellaSwag, and MMLU. Internet-trained models are trained in a low-bandwidth setting, while centralized-cluster models are trained using conventional large-scale datacenter infrastructure and AdamW.}
\label{tab:internet-vs-centralized}
\begin{tabular}{l l c c c c c c}
\toprule
\textbf{Model} & \textbf{Compute Env.} & \textbf{Size} & \textbf{Tokens} & \textbf{ARC-C} & \textbf{ARC-E} & \textbf{HellaSwag} & \textbf{MMLU} \\
\midrule
Intellect-1 & Internet & 10B & 1T & 44.8 & 71.6 & 70.5 & 32.7 \\
Ours  & Internet & 72B & 1T & 56.48 & 81.73 & 78.00 & 61.09 \\
\midrule
LLM360 K2 & Datacenter & 65B & 1.4T & 53.84 & 75.93 & 82.81 & 63.90 \\
LLaMA-2-7B & Datacenter & 7B & 2T & 45.90 & 74.58 & 75.92 & 40.86 \\
LLaMA-2-70B & Datacenter & 70B & 2T & 57.59 & 80.77 & 83.86 & 65.56 \\
\bottomrule
\end{tabular}
\end{table}

\section{Mixture-of-Experts Experiments}
\label{app:moe}

While our main experiments use dense models, SparseLoCo can be applied to MoE transformers without algorithmic changes. In this section, we evaluate SparseLoCo against baselines using a Qwen2-MoE-style architecture~\citep{qwen2}. The model has 645M total parameters and 273M active parameters per token (645M-A273M). The model uses 12 decoder layers, hidden size 1024, 8 attention heads, 8 key-value heads, and sequence length 2048, with an MoE FFN in every decoder layer. We use 32 routed experts with top-4 routing and routed-expert intermediate size 358 (granularity $1/8$), and one shared expert with intermediate size 2816. The router auxiliary loss coefficient is 0.001.

We sweep AdamW over learning rate
\vspace{0.8mm}$\{10^{-3}, \mathbf{3{\times}10^{-3}}, 5{\times}10^{-3}\}$ and
$\beta_2\in\{0.9,\mathbf{0.95},0.99\}$, and DiLoCo over
$\alpha_{\text{inner}}\in\{10^{-3}, \mathbf{8{\times}10^{-4}}, 6{\times}10^{-4}\}$ and
$\alpha_{\text{outer}}\in\{\mathbf{0.6},0.8,1.0\}$. We did not run an MoE-specific sweep for SparseLoCo and instead reused the dense-model hyperparameters from Appendix~\ref{app:sweeps}. As shown in Table~\ref{tab:moe_results}, SparseLoCo nearly matches AdamW at $3.12\%$ density and clearly outperforms DiLoCo.

\section{Wall-clock Time Under Bandwidth Constraints}
\label{app:wallclock}

\begin{figure}[H]
\centering
\includegraphics[width=0.65\columnwidth]{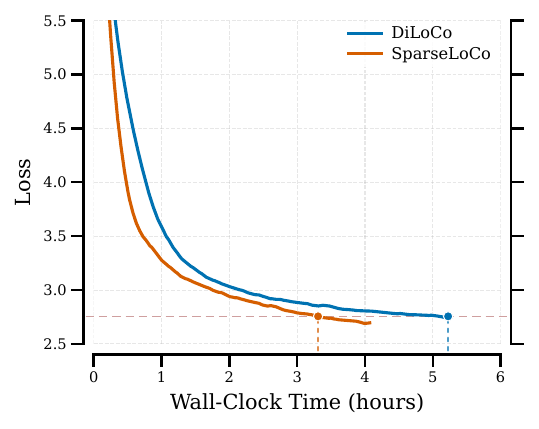}
\caption{\textbf{Wall-clock improvement under a 300\,Mbps bandwidth limit.}
Validation loss vs.\ wall-clock time for a 512M model trained with $H{=}15$, $R{=}8$ workers, and a 10B-token budget. SparseLoCo reaches DiLoCo's final loss roughly two hours earlier.}
\label{fig:wallclock}
\end{figure}

We measure wall-clock training time in our main 512M setting ($R{=}8$ workers, $H{=}15$, 10B-token budget, using $8\times$H100 GPUs) under a realistic $\sim$300\,Mbps cross-replica bandwidth constraint (a common setting for machines rented from vast.ai). Figure~\ref{fig:wallclock} plots validation loss against wall-clock time. SparseLoCo hits DiLoCo's final loss about two hours earlier. We also observed that the overhead of SparseLoCo's compression ($\textsc{Top-}k$, quantization, chunking) is negligible: only $0.03\%$ extra compute per outer step. Finally, as the pseudo-gradient size scales with model size, we expect the wall-clock gap to increase with larger models at the same bandwidth.

\section{Compute Utilization vs.\ Bandwidth}
\label{app:compute-util}
\begin{figure}[t]
    \centering
    \includegraphics[width=0.97\linewidth]{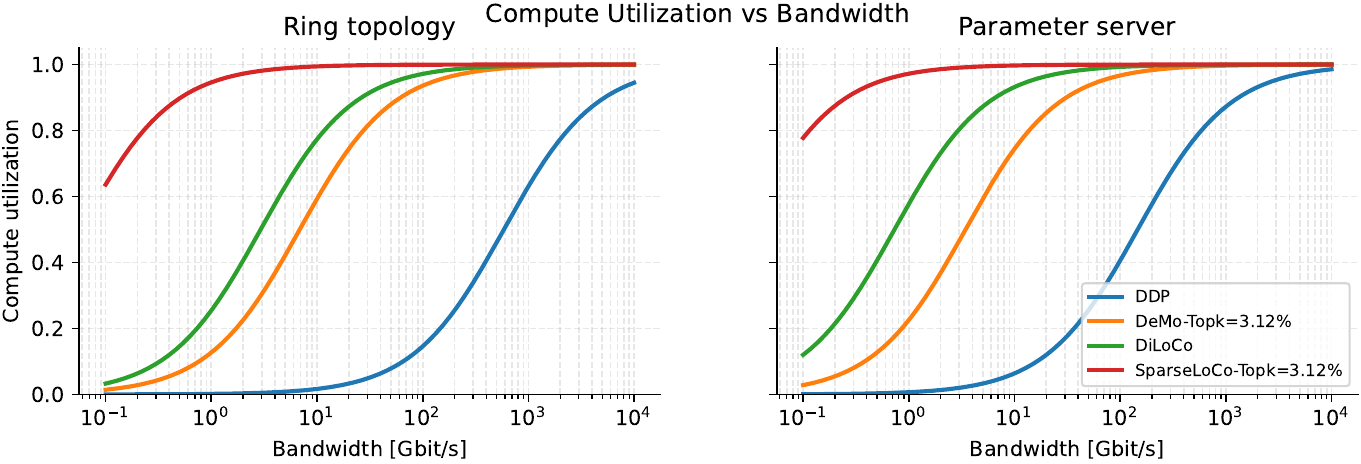}
    \caption{\rev{Compute utilization vs.\ bandwidth for DDP, DeMo, DiLoCo, and SparseLoCo under ring (left) and parameter-server (right) topologies. Compute utilization is calculated as total time spent in computation over full training time (including communication). We simulate a 70B LLaMA-2 model trained with $R{=}16$ replicas each with $8\times$ B200, assuming a reasonable 40\% MFU under different bandwidth settings.}}
    \label{fig:compute-util}
\end{figure}

We estimate the compute utilization of each method as \(\frac{T_{\text{compute}}}{T_{\text{compute}} + T_{\text{comms}}}\), where \(T_{\text{compute}}\) is the time spent in pure computation and \(T_{\text{comms}}\) is the time spent in communication. We first estimate \(T_{\text{compute}}\) considering the FLOPs profile of the model, assuming 8$\times$B200 GPUs per worker with $R{=}16$ workers, a theoretical FP16/BF16 throughput of $4.5\times 10^{15}$ FLOPs/s per GPU, and a reasonable machine FLOP utilization (MFU) of 40\%. Then, we simulate training calculating \(T_{\text{comms}}\) under different bandwidth constraints considering the pseudo-gradient message sizes of each method (Figure~\ref{fig:compute-util}). At low bandwidths, SparseLoCo achieves substantially higher utilization than DDP, DeMo, and DiLoCo. For instance, at $1$\,Gbit/s, SparseLoCo exceeds 95\% utilization, significantly outperforming the baselines.

\section{Effect of Chunking, DCT, and inner steps}\label{sec:chunking}
A recently introduced method~\cite{peng2024decoupled} considered the single-step setting with error feedback, using a compression function that first applies a discrete cosine transform (DCT) on tensor chunks and then selects the $\textsc{Top-}k$ values in the DCT domain. It further employed sign descent on the final aggregated update~\cite{kunstner2023noise}. Without the DCT transform, this approach can be seen as a special case of SparseLoCo when $H=1$, the inner optimizer is plain SGD, and the outer optimizer utilizes sign descent. Since the effect of the DCT transform, designed for data with sequential structure where the order of elements matters, is not well understood in this context, and given the additional uncertainty about the role of chunking, in this section we disentangle the contributions of both for DeMo, as well as in the multi-step ($H > 1$) setting. The ablations of these three factors, evaluated purely in terms of loss, are presented in Table~\ref{tab:ablate2}. Here, the $\textsc{Top-}k$ EF baseline is a simplified DeMo that applies $\textsc{Top-}k$ selection globally to the entire tensor (rather than within chunks) while still utilizing sign descent.

We observe that in the setting with no local steps ($\textsc{Top-}k$ EF, DeMo) the impact of chunking is very significant and the performance of DeMo can be nearly recovered without resorting to the DCT. When using the local setting ($H{>}1$), we observed that DCT actually degrades performance; however, we also find that the impact of chunking is more limited than in the setting of $H{=}1$. We hypothesize that chunking and DCT both serve to reduce the effect of outlier values on the scale of individual workers' contributions, which may be less critical in the case of SparseLoCo due to its adaptive inner optimizer. 

Notably, DeMo does not have a natural way to incorporate adaptive optimization, and in practice, the sign descent is used to approximate the benefits of the Adam optimizer~\cite{peng2024decoupled}. 
A significant advantage of SparseLoCo is that operating on the pseudo-gradients allows easy integration of adaptive optimizers like Adam in the inner loop.

\begin{table}[H]
\centering
\caption{Ablation of tensor chunking and DCT (lower loss is better). We observe that chunking is critical for the performance of DeMo. With $H{>}1$, DCT degrades performance. All runs use full precision (FP32).}
\label{tab:ablate2}
\begin{tabular}{lcc}
\toprule
\textbf{Method} & \textbf{No DCT} & \textbf{DCT} \\
\midrule
SparseLoCo ($H > 1$, Chunking, $\textsc{Top-}k$ EF)           & \textbf{2.72} & 2.75 \\
SparseLoCo w/o Chunking ($H > 1$, $\textsc{Top-}k$ EF)        & 2.73 & 2.76 \\
DeMo (Chunking, $\textsc{Top-}k$ EF w/ Sign Descent)          & 2.87 & 2.83 \\
$\textsc{Top-}k$ EF (w/ Sign Descent)                         & 3.48 & 2.84 \\
DiLoCo                                               & 2.76 & -- \\
\bottomrule
\end{tabular}
\end{table}

\section{Streaming SparseLoCo}\label{sec:streaming}
In this section, we verify that Streaming DiLoCo~\cite{streamingdiloco} and SparseLoCo can be combined. Streaming DiLoCo is an orthogonal direction to SparseLoCo for reducing peak communication volume by hiding it. SparseLoCo reduces the absolute number of bits communicated per step through compression, thus indirectly reducing peak communication. Streaming DiLoCo directly reduces peak bandwidth by only communicating subsets of the model's parameters at a time but does not reduce the absolute number of bits communicated.

Table~\ref{tab:streamingspraseloco} reports results for combining SparseLoCo with Streaming DiLoCo to reduce peak communication volume. We train 18-layer 1B-parameter transformers (hidden dimension $2048$) in this ablation. The model is partitioned into three even subsets of $6$ hidden layers, with the first and third subsets containing the embedding and unembedding layers, respectively. We train the $1,055$M parameter model for a Chinchilla-optimal $21$B tokens~\cite{hoffmann2022training}. We use a communication interval of $H=15$ for the full model (Streaming communicates every $5$ steps) and $8$ workers.  We observe that both models reach the same final validation loss (it differed only in the 4th decimal), while Streaming SparseLoCo reduces peak communication volume by a factor of $3$.

\begin{table}[H]
\centering
\caption{We combine SparseLoCo with Streaming DiLoCo to reduce peak communication volume when training an 18-layer 1B parameter transformer. The model is partitioned into three even subsets of $6$ hidden layers, with the first and third subsets containing the embedding and unembedding layers, respectively. We use a communication interval of $H=15$ for the full model (Streaming communicates every $5$ steps) and $8$ workers. We observe that both models reach the same final validation loss (it differed only in the 4th decimal), while Streaming SparseLoCo reduces peak communication volume by a factor of $3$. }
\label{tab:streamingspraseloco}
\begin{tabular}{lcccc}
\toprule
\textbf{Method}     & \textbf{Density} & \textbf{Comm. Volume/Step}& \textbf{Peak Comm. Volume} & \textbf{Loss} \\
\midrule
SparseLoCo & 3.125\%    & 35.03 MB&35.03 MB&\textbf{2.51} \\
Streaming SparseLoCo & 3.125\%   & 35.03 MB&11.68 MB&\textbf{2.51} \\
\bottomrule
\end{tabular}
\end{table}

\section{Freezing Error Feedback}\label{app:ef-freeze}
We apply a short error feedback (OuterEF) freeze at the beginning of training: for the first few outer steps, the error feedback $e_r$ is not utilized. Concretely, during the freeze we don’t use nor accumulate in the EF buffer. We find that freezing the OuterEF for the first few outer steps slightly improves training stability and overall performance (see Table~\ref{tab:ef-freeze}).

\begin{table}[H]
\centering
\caption{Freezing error feedback for the first few outer steps improves training. The final validation loss for 512M models trained with SparseLoCo (3.12\% density), $R{=}8$ replicas, and communication interval $H{=}15$ is reported.}
\label{tab:ef-freeze}
\begin{tabular}{lcc}
\toprule
\textbf{EF Freeze} & \textbf{Loss} \\
\midrule
0\% & 2.704\\
5\% & 2.699\\
\bottomrule
\end{tabular}
\end{table}

\section{Error Feedback Decay}
\label{app:ef-decay}

We sweep the error feedback decay coefficient $\beta\in\{0.9, 0.95, 0.999\}$ using all other SparseLoCo hyperparameters from Appendix~\ref{app:sweeps} across different settings. Table~\ref{tab:beta_sensitivity} reports results for 178M and 512M models at $0.78\%$ and $3.12\%$ density, with $H\in\{15,50\}$. We find that $\beta{=}0.95$ is a stable choice across all settings.

\begin{table}[H]
\centering
\caption{\textbf{SparseLoCo is robust to the error feedback decay $\beta$.}
Final validation loss for different $\beta$ values, with the remaining SparseLoCo hyperparameters fixed as in Appendix~\ref{app:sweeps}. Best result in each row is bolded.}
\label{tab:beta_sensitivity}
\begin{tabular}{lccccc}
\toprule
Model & Density & $H$ & $\beta{=}0.9$ & $\beta{=}0.95$ & $\beta{=}0.999$ \\
\midrule
178M & 3.12\% & 50 & \textbf{2.99} & \textbf{2.99} & \textbf{2.99} \\
178M & 0.78\% & 50 & 3.09 & \textbf{3.08} & 3.21 \\
512M & 3.12\% & 15 & 2.71 & \textbf{2.70} & 2.72 \\
\bottomrule
\end{tabular}
\end{table}

\section{Compression of indices in $\textsc{Top-}k$}\label{app:compression_size}
In $\textsc{Top-}k$ methods, the indices of the selected values need to be transmitted alongside the values. When values are aggressively quantized (as in SparseLoCo), this index-transmission overhead becomes significant. In SparseLoCo, we utilize chunk sizes of $C{=}4096$, so, naively, we can transmit indices in $12$ bits per transmitted value. However, with 2-bit quantization, this overhead becomes significant, motivating further index compression. Assuming a chunk size of $C$ and $\textsc{Top-}k$ selection, we observe that the information-theoretic limit is $\log_2\!\binom{C}{k}$ bits. For practical cases considered in this work ($C{=}4096$ and $k\in\{32,128,256\}$), this corresponds to 8.3, 6.3, and 5.3 bits per transmitted value, respectively. In practice, we designed a custom compression algorithm based on sub-chunking and Rice coding that achieves 8.9, 6.6, and 5.6 bits per value for these cases, with negligible runtime overhead.

\section{Scaling Replicas Across Different Densities and Communication Intervals}\label{app:sacling-R}
We compare DiLoCo and SparseLoCo while varying the number of workers $R\in\{8,16,32\}$, communication intervals $H\in\{15,50,100\}$ using model sizes 178M and 512M, and report the final validation loss in Tables~\ref{tab:178M_scaling_r}, \ref{tab:512M_scaling_r}, and \ref{tab:workers_H50}. We observe that SparseLoCo outperforms DiLoCo with higher number of parallel workers.%

\begin{table}[H]
\caption{Final validation loss for the 178M model while varying the number of workers ($R\in\{8,16,32\}$) and the communication interval ($H\in\{15,50,100\}$). \textbf{Best} is bold.}
\label{tab:178M_scaling_r}
\centering

\noindent
\begin{minipage}[t]{0.48\linewidth}
\centering
\textbf{H=15}\\[0.25em]
\begin{tabular}{lccc}
\toprule
Method & Density & \multicolumn{2}{c}{Loss} \\
\cmidrule(lr){3-4}
 &  & R=8 & R=32 \\
\midrule
AdamW   & 100.00\% & 2.91 & 2.91 \\
DiLoCo  & 100.00\% & 2.97 & 3.10 \\
\midrule
\multirow{7}{*}{\footnotesize SparseLoCo}
 & 0.78\%  & 2.96 & 3.02 \\
 & 1.56\%  & 2.93 & 3.00 \\
 & 3.12\%  & \textbf{2.91} & \textbf{2.99} \\
 & 6.25\%  & 2.94 & 3.00 \\
 & 12.50\% & 2.96 & 3.04 \\
 & 25.00\% & 2.96 & 3.14 \\
 & 50.00\% & 2.99 & 3.29 \\
\bottomrule
\end{tabular}
\end{minipage}
\hfill%
\begin{minipage}[t]{0.48\linewidth}
\centering
\textbf{H=50}\\[0.25em]
\begin{tabular}{lccc}
\toprule
Method & Density & \multicolumn{2}{c}{Loss} \\
\cmidrule(lr){3-4}
 &  & R=8 & R=32 \\
\midrule
AdamW   & 100.00\% & 2.91 & 2.91 \\
DiLoCo  & 100.00\% & 3.05 & 3.20 \\
\midrule
\multirow{7}{*}{\footnotesize SparseLoCo}
 & 0.78\%  & 3.09 & 3.13 \\
 & 1.56\%  & 3.03 & \textbf{3.07} \\
 & 3.12\%  & \textbf{2.99} & 3.09 \\
 & 6.25\%  & \textbf{2.98} & 3.09 \\
 & 12.50\% & 3.00 & 3.14 \\
 & 25.00\% & 3.04 & 3.25 \\
 & 50.00\% & 3.12 & 3.42 \\
\bottomrule
\end{tabular}
\end{minipage}

\vspace{0.8\baselineskip}

\noindent\hfill
\begin{minipage}[t]{0.48\linewidth}
\centering
\textbf{H=100}\\[0.25em]
\begin{tabular}{lccc}
\toprule
Method & Density & \multicolumn{2}{c}{Loss} \\
\cmidrule(lr){3-4}
 &  & R=8 & R=32 \\
\midrule
AdamW & 100.00\% & 2.91 & 2.91 \\
DiLoCo & 100.00\% & 3.12 & 3.29 \\
\midrule
\multirow{7}{*}{SparseLoCo}
 & 0.78\%  & 3.20 & 3.29 \\
 & 1.56\%  & 3.12 & 3.26 \\
 & 3.12\%  & 3.05 & 3.19 \\
 & 6.25\%  & \textbf{3.03} & \textbf{3.17} \\
 & 12.50\% & \textbf{3.03} & 3.21 \\
 & 25.00\% & 3.07 & 3.32 \\
 & 50.00\% & 3.17 & 3.48 \\
\bottomrule
\end{tabular}
\end{minipage}\hfill\mbox{}

\end{table}

\begin{table}[t]
\caption{Final evaluation loss of scaling number of workers $R\in\{8,16,32\}$ for different communication interval $H\in\{15,100\}$ using different communication densities for SparseLoCo using 512M model size. Best results in each communication interval are presented in \textbf{Bold}.}
\label{tab:512M_scaling_r}
\centering
\begin{tabular}{lcccc}
\toprule
Method & Density & Loss (R=8) & Loss (R=16) & Loss (R=32) \\
\midrule
AdamW & 100.00\% & 2.69 & 2.69 & 2.69 \\
\midrule

&& \multicolumn{3}{c}{H=15} \\
\cmidrule(lr){3-5} 
DiLoCo & 100.00\% & 2.76 & 2.77 & 2.82 \\

\noalign{\vskip 3pt}\cdashline{2-5}\noalign{\vskip 3pt}%

\multirow{6}{*}{SparseLoCo}
 & 0.78\%  & 2.79 & 2.81 & 2.84 \\
 & 1.56\%  & 2.74 & 2.76 & 2.79 \\
 & 3.12\%  & \textbf{2.70} & \textbf{2.74} & \textbf{2.77} \\
 & 6.25\%  & 2.71 & 2.76 & 2.78 \\
 & 12.50\% & 2.76 & 2.78 & 2.82 \\
 & 25.00\% & 2.77 & 2.78 & 2.93 \\

\midrule
&& \multicolumn{3}{c}{H=100} \\
\cmidrule(lr){3-5} 
DiLoCo & 100.00\% & 2.87 & 2.94 & 3.05 \\
\noalign{\vskip 3pt}\cdashline{2-5}\noalign{\vskip 3pt}%
\vspace{0.4mm}
\multirow{6}{*}{SparseLoCo}
 & 0.78\%  & 3.09 & 3.14 & 3.29 \\
 & 1.56\%  & 3.02 & 3.06 & 3.21 \\
 & 3.12\%  & 2.96 & 3.03 & 3.12 \\
 & 6.25\%  & 2.94 & 2.97 & 3.03 \\
 & 12.50\% & 2.85 & \textbf{2.88} & \textbf{3.02} \\
 & 25.00\% & \textbf{2.82} & 2.89 & 3.11 \\

\bottomrule
\end{tabular}
\end{table}

\begin{table}[t]

\caption{Model settings for 178M (left) and 2B (right) model scales.}
\label{tab:model-hparams-178M2B}

\noindent
\begin{minipage}[t]{0.49\linewidth}
\centering
\begin{tabular}{ll}
\toprule
\textbf{Parameter} & \textbf{Value} \\
\midrule
Total Parameters  & 177,622,016 \\
Number of Layers  & 9 \\
Hidden Size       & 1,024 \\
Intermediate Size & 2,688 \\
Attention Heads   & 8 \\
Vocabulary Size   & 32,000 \\
FFN Activation    & SwiGLU \\
\bottomrule
\end{tabular}
\end{minipage}%
\begin{minipage}[t]{0.49\linewidth}
\centering
\begin{tabular}{ll}
\toprule
\textbf{Parameter} & \textbf{Value} \\
\midrule
Total Parameters  & 1,972,759,040 \\
Number of Layers  & 24 \\
Hidden Size       & 2,560 \\
Intermediate Size & 7,680 \\
Attention Heads   & 20 \\
Key-Value Heads   & 5 \\
Vocabulary Size   & 32,000 \\
FFN Activation    & SwiGLU \\
\bottomrule
\end{tabular}
\end{minipage}

\end{table}

\section{Overtraining Regime with Large Communication Interval}
Following~\cite{scalingdiloco}, we put SparseLoCo to the test in an overtraining regime by doubling the token budget to 20B and using a larger communication interval of $H{=}250$. Our observations are consistent with the trends in Figure~\ref{fig:comm-freq}, and SparseLoCo outperforms \textsc{DiLoCo} at this setting (Table~\ref{tab:ablate-overtraining}).

\begin{table}[H]
\centering
\caption{Overtraining on $2\times$ data (20B token budget) with communication interval $H{=}250$.}
\label{tab:ablate-overtraining}
\begin{tabular}{lcc}
\toprule
\textbf{Method}     & \textbf{Density} & \textbf{Loss} \\
\midrule
DiLoCo     & 100\%   & 2.77 \\
SparseLoCo & 50\%    & \textbf{2.73} \\
SparseLoCo & 25\%    & 2.74 \\
SparseLoCo & 12.5\%  & 2.79 \\
SparseLoCo & 3.12\%  & 2.97 \\
SparseLoCo & 1.56\%  & 3.00 \\
\bottomrule
\end{tabular}
\end{table}

\begin{revblock}
\section{GPT-2 Experiments}
\label{app:gpt2}

To verify that SparseLoCo applies beyond LLaMA-style models, we also evaluate a 512M-parameter GPT-2 model. We reuse the best hyperparameters from the 512M LLaMA setting for both DiLoCo and SparseLoCo and train with $H{=}15$ inner steps on the same dataset and token budget.
\end{revblock}

\begin{table}[H]
\centering
\caption{\rev{\textbf{SparseLoCo vs.\ DiLoCo on a 512M-parameter GPT-2 model.} We report the final validation loss.}}
\label{tab:gpt2_results}
\begin{tabular}{lcc}
\toprule
Method & Density & Loss \\
\midrule
SparseLoCo & 3.12\% & 2.89 \\
DiLoCo     & 100\%  & 2.92 \\
\bottomrule
\end{tabular}
\end{table}

\section{Equivalence of LOM and Global Outer Momentum}\label{app:lom-equivalence}
We show that DiLoCo-LOM iterates are actually equivalent to DiLoCo.
\begin{proposition}\label{prop:lom_equals_global}

Suppose identical initialization $m_r^{(0)}=m^{(0)}=0$ for all $r\in[R]$, and fixed outer-momentum coefficient $\beta\in[0,1)$. Then, for all $t\ge 0$,
\[
\bar m^{(t)} = m^{(t)} \quad\text{and}\quad \bar{\tilde{\Delta}}^{(t)} = \tilde{\Delta}^{(t)} ,
\]
where $\bar{m}^{(t)} := \tfrac{1}{R}\sum_{r=1}^R m_r^{(t)}$ denotes the average of local momentum buffers, $\bar{\tilde{\Delta}}^{(t)} := \tfrac{1}{R}\sum_{r=1}^R \tilde{\Delta}_r^{(t)}$ the averaged LOM Nesterov direction, and $m^{(t)}$ and $\tilde{\Delta}^{(t)}$ the global momentum and Nesterov direction in DiLoCo, respectively. Consequently, the parameter updates of DiLoCo-LOM and DiLoCo are identical at every time step.
\end{proposition}

{\renewcommand{\qedsymbol}{}\begin{proof}
We first show $\bar m^{(t)}=m^{(t)}$ by induction, then obtain $\bar{\tilde{\Delta}}^{(t)} = \tilde\Delta^{(t)}$ by linearity.

\emph{Base case ($t=0$).} With $m_r^{(0)}=0$ and $m^{(0)}=0$, we have $\bar m^{(0)}=\tfrac{1}{R}\sum_r m_r^{(0)}=0=m^{(0)}$.

\emph{Inductive step.} Assume $\bar m^{(t-1)}=m^{(t-1)}$ for some $t\ge 1$. Averaging the local recursion,
\[
\bar m^{(t)}
= \tfrac{1}{R}\sum_{r=1}^R \big(\beta\, m_r^{(t-1)} + \Delta_r^{(t)}\big)
= \beta \big(\tfrac{1}{R}\sum_r m_r^{(t-1)}\big) + \tfrac{1}{R}\sum_r \Delta_r^{(t)}
= \beta\, \bar m^{(t-1)} + \bar\Delta^{(t)}.
\]
By the global recursion, $m^{(t)}=\beta m^{(t-1)}+\bar\Delta^{(t)}$, hence $\bar m^{(t)}=m^{(t)}$. For the Nesterov directions,
\[
\bar{\tilde{\Delta}}^{(t)}
= \tfrac{1}{R}\sum_{r=1}^R \big(\Delta_r^{(t)} + \beta\, m_r^{(t)}\big)
= \bar\Delta^{(t)} + \beta \big(\tfrac{1}{R}\sum_r m_r^{(t)}\big)
= \bar\Delta^{(t)} + \beta\, \bar m^{(t)}
= \bar\Delta^{(t)} + \beta\, m^{(t)}
= \tilde\Delta^{(t)}.\qedhere
\]
\end{proof}}    

\begin{revblock}
\section{Sparsity Ablations for DeMo and SparseLoCo}
\label{app:demo-density}

In \figurename~\ref{fig:demo-density}, we compare DeMo and SparseLoCo at different sparsity levels with DiLoCo, and DiLoCo without outer momentum. We train 512M models with $R{=}8$ replicas, and for the multi-step baselines, we use a fixed communication interval of $H{=}15$.

\begin{figure}[H]
    \centering

    \includegraphics[width=0.65\linewidth]{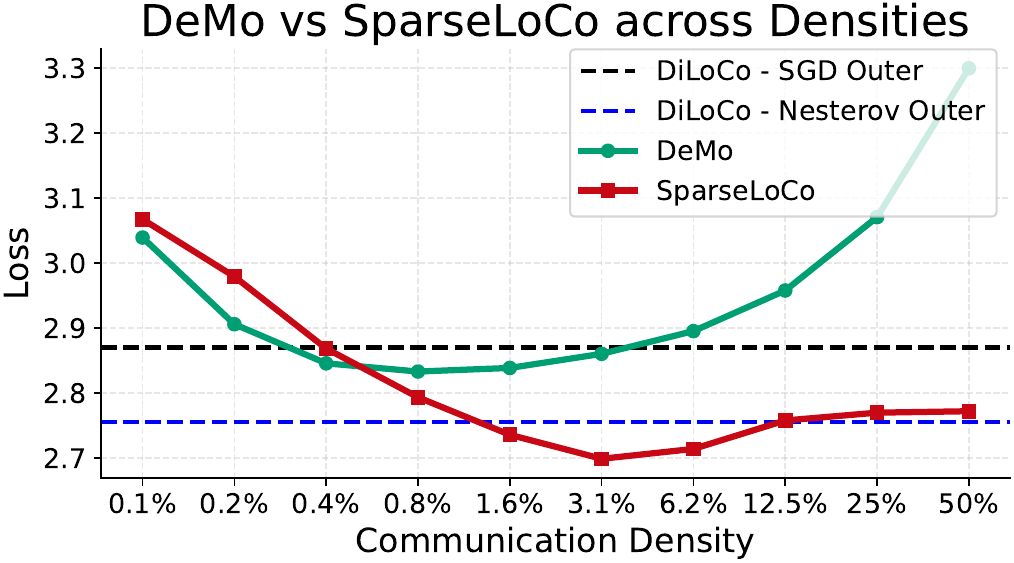}

    \caption{\rev{\textbf{DeMo and SparseLoCo across varying communication densities.} We compare DeMo and SparseLoCo across varying sparsity levels using the same settings as Figure~\ref{fig:comm-freq}; multi-step methods use a communication interval of $H{=}15$.}}
    \label{fig:demo-density}
\end{figure}

\end{revblock}

\section{Exact algorithmic description of DiLoCo variants}
\label{app:variant-definitions}

In this appendix we summarize the DiLoCo variants introduced in this work.
All variants share the same local inner loop as DiLoCo: each worker starts from the synchronized parameters
$\theta^{(t-1)}$, runs $H$ local AdamW steps, and forms the pseudo-gradient
\[
\Delta_r^{(t)}=\theta^{(t-1)}-\theta_r^{(t)}.
\]
These methods differ only
in the outer step; below we define each variant and provide the corresponding outer-step algorithms. Note that all $\textsc{Top-}k$ operations are applied chunk-wise as in SparseLoCo.

\paragraph{DiLoCo-LOM.}
DiLoCo-LOM replaces the global DiLoCo outer momentum with a local outer momentum buffer
$m_r$ on each worker. Each worker updates its local momentum using its own pseudo-gradient
$\Delta_r^{(t)}$ and forms a local Nesterov direction $\tilde{\Delta}_r^{(t)}$. These directions are
then communicated and aggregated across workers to advance the model. Although the momentum
buffers are local, the resulting update is exactly equivalent to standard DiLoCo by linearity, as shown
in Appendix~\ref{app:lom-equivalence}.

\paragraph{DiLoCo-LOM-Sub-$k$.}
DiLoCo-LOM-Sub-$k$ uses the same dense DiLoCo-LOM update, but after the outer update it
subtracts the largest-magnitude entries of each local momentum buffer (Line~\ref{line:lom-subk} in Algorithm~\ref{alg:lom-subk}). This creates a bridge between DiLoCo
and SparseLoCo: communication remains dense as in DiLoCo, while the momentum buffer behaves like an EF residual by removing the entries that would have been selected by $\textsc{Top-}k$ after
each outer step.

\paragraph{DiLoCo+\textsc{Top-}$k$ with optional EF.}
DiLoCo+\textsc{Top-}$k$ compresses each local pseudo-gradient before communication, then applies the standard DiLoCo Nesterov outer optimizer to the aggregated sparse pseudo-gradients. Without EF, workers directly sparsify $\Delta_r^{(t)}$. Workers may also incorporate an optional local EF buffer (Line~\ref{line:topk-before-ef} of Algorithm~\ref{alg:topk-before-momentum}). This is the most direct \textsc{Top-}$k$-compressed DiLoCo baseline because only the communicated pseudo-gradient is changed, while the DiLoCo outer optimizer remains unchanged.

\paragraph{DiLoCo-LOM+\textsc{Top-}$k$ with optional EF.}
DiLoCo-LOM+\textsc{Top-}$k$ applies compression after local outer momentum. Each worker first updates its local momentum, forms a local Nesterov direction, and then sparsifies that direction before communication. Workers may also incorporate an optional local EF buffer (Line~\ref{line:topk-after-ef} of Algorithm~\ref{alg:topk-after-lom}). This provides another natural baseline for combining sparsification with DiLoCo-style outer momentum, discussed in Appendix~\ref{app:lom-topk-ef-ablation}.

\begin{algorithm}[H]
\caption{Outer step for DiLoCo-LOM and DiLoCo-LOM-Sub-$k$}
\label{alg:lom-subk}
\small
\begin{algorithmic}[1]
  \Require Pseudo-gradients $\{\Delta_r^{(t)}\}_{r=1}^R$; local momentum buffers
  $\{m_r^{(t-1)}\}_{r=1}^R$; outer learning rate $\alpha$; momentum coefficient $\mu$.

  \ForAll{$r\in[R]$}
    \State $m_r^{(t)} \gets \mu m_r^{(t-1)}+\Delta_r^{(t)}$
    \Comment{local momenta}
    \State $\tilde{\Delta}_r^{(t)} \gets \Delta_r^{(t)}+\mu m_r^{(t)}$
    \Comment{local Nesterov directions}
  \Statex

  \State $\tilde{\Delta}^{(t)} \gets \frac{1}{R}\sum_{r=1}^R \tilde{\Delta}_r^{(t)}$
  \Comment{aggregate local Nesterov directions}
  \State $\theta^{(t)} \gets \theta^{(t-1)}-\alpha \tilde{\Delta}^{(t)}$

  \Statex
  \State \algvariantbox{blue}{%
    \textbf{If DiLoCo-LOM-Sub-$k$:} \hspace*{13.5em}\textcolor{gray}{\footnotesize // Sub-$k$: remove top-$k$ values from local momenta}\\
    \hspace*{1.5em}$m_r^{(t)} \gets m_r^{(t)}-\textsc{Top-}k(m_r^{(t)})$
    \label{line:lom-subk}
  }\vspace{1mm}
  \EndFor
\end{algorithmic}
\end{algorithm}

\begin{algorithm}[H]
\caption{Outer step for DiLoCo+\textsc{Top-}$k$[+EF]}
\label{alg:topk-before-momentum}
\small
\begin{algorithmic}[1]
  \Require Pseudo-gradients $\{\Delta_r^{(t)}\}_{r=1}^R$; global momentum $m^{(t-1)}$;
  optional EF buffers $\{e_r^{(t-1)}\}_{r=1}^R$; outer learning rate $\alpha$;
  momentum coefficient $\mu$; EF decay $\rho$.

  \ForAll{$r\in[R]$}
    \If{use EF} \Comment{Optional EF}
      \Statex \label{line:topk-before-ef}\algvariantbox{purple}{%
        \hspace*{3em}$u_r^{(t)} \gets \rho e_r^{(t-1)}+\Delta_r^{(t)}$\\
        \hspace*{3em}$\hat{\Delta}_r^{(t)} \gets \textsc{Top-}k(u_r^{(t)})$\\
        \hspace*{3em}$e_r^{(t)} \gets u_r^{(t)}-\hat{\Delta}_r^{(t)}$
      }
    \Else
      \State $\hat{\Delta}_r^{(t)} \gets \textsc{Top-}k(\Delta_r^{(t)})$ \Comment{No EF: sparsify the local pseudo-gradients directly}
    \EndIf

    \Statex
  \State $\Delta^{(t)} \gets \frac{1}{R}\sum_{r=1}^R \hat{\Delta}_r^{(t)}$
  \Comment{aggregate sparse pseudo-gradients}
  \State $m^{(t)} \gets \mu m^{(t-1)}+\Delta^{(t)}$
  \Comment{standard DiLoCo momentum}
  \State $\tilde{\Delta}^{(t)} \gets \Delta^{(t)}+\mu m^{(t)}$
  \Comment{standard DiLoCo Nesterov direction}
  \State $\theta^{(t)} \gets \theta^{(t-1)}-\alpha\tilde{\Delta}^{(t)}$
  \EndFor
\end{algorithmic}
\end{algorithm}

\begin{algorithm}[H]
\caption{Outer step for DiLoCo-LOM+\textsc{Top-}$k$[+EF]}
\label{alg:topk-after-lom}
\small
\begin{algorithmic}[1]
  \Require Pseudo-gradients $\{\Delta_r^{(t)}\}_{r=1}^R$; local momentum buffers
  $\{m_r^{(t-1)}\}_{r=1}^R$; optional EF buffers $\{e_r^{(t-1)}\}_{r=1}^R$;
  outer learning rate $\alpha$; momentum coefficient $\mu$; EF decay $\rho$.

  \ForAll{$r\in[R]$}
    \State $m_r^{(t)} \gets \mu m_r^{(t-1)}+\Delta_r^{(t)}$
    \Comment{local momenta}
    \State $\tilde{\Delta}_r^{(t)} \gets \Delta_r^{(t)}+\mu m_r^{(t)}$
    \Comment{local Nesterov directions}
    \Statex
    \If{use EF} \Comment{Optional EF}
      \Statex \label{line:topk-after-ef}\algvariantbox{purple}{%
        \hspace*{3em}$u_r^{(t)} \gets \rho e_r^{(t-1)}+\tilde{\Delta}_r^{(t)}$\\
        \hspace*{3em}$\hat{\Delta}_r^{(t)} \gets \textsc{Top-}k(u_r^{(t)})$\\
        \hspace*{3em}$e_r^{(t)} \gets u_r^{(t)}-\hat{\Delta}_r^{(t)}$
      }
    \Else
      \State $\hat{\Delta}_r^{(t)} \gets \textsc{Top-}k(\tilde{\Delta}_r^{(t)})$
        \Comment{No EF: sparsify the local Nesterov direction directly.}
    \EndIf

    \Statex
  \State $\Delta^{(t)} \gets \frac{1}{R}\sum_{r=1}^R \hat{\Delta}_r^{(t)}$
  \Comment{aggregate compressed local directions}
  \State $\theta^{(t)} \gets \theta^{(t-1)}-\alpha\Delta^{(t)}$
\EndFor
\end{algorithmic}
\end{algorithm}

\clearpage
\section{DiLoCo-LOM with \textsc{Top-}$k$ and EF}
\label{app:lom-topk-ef-ablation}

In Sections~\ref{sec:topk-ef-localadam} and~\ref{sec:naive-topk-diloco}, we study the naive and natural combination of DiLoCo and \textsc{Top-}$k$+EF sparsification, where we apply the \textsc{Top-}$k$ operation, optionally with EF, to the local pseudo-gradients before communication occurs in the original DiLoCo algorithm~\cite{douillard2023diloco}. This means that the outer momentum is updated from the same aggregated signal on all replicas, so the momentum state remains synchronized across replicas. Since DiLoCo-LOM is equivalent to DiLoCo in the dense case (see Appendix~\ref{app:lom-equivalence} for the proof), another way to combine \textsc{Top-}$k$ and DiLoCo is to reverse this ordering: first form the local Nesterov directions as in DiLoCo-LOM, and then apply \textsc{Top-}$k$+EF to those directions before communication. We refer to this variant as DiLoCo-LOM+\textsc{Top-}$k$+EF.

We establish the performance of DiLoCo-LOM+\textsc{Top-}$k$+EF on a 178M model with $R{=}8$, $H{=}15$, and $3.12\%$ density. As shown in Table~\ref{tab:lom_topk_ef}, it improves over DiLoCo+\textsc{Top-}$k$+EF, but still underperforms both dense DiLoCo and SparseLoCo. These results demonstrate that naive combinations of \textsc{Top-}$k$ + EF with DiLoCo and with DiLoCo-LOM are outperformed by SparseLoCo.

\begin{table}[H]
\centering
\caption{\textbf{Additional DiLoCo-LOM+\textsc{Top-}$k$+EF ablation.}
We report final validation loss for 178M models with $R{=}8$ replicas and communication interval
$H{=}15$. Sparse methods use $3.12\%$ density.}
\label{tab:lom_topk_ef}
\begin{tabular}{lcc}
\toprule
Method & Density & Loss \\
\midrule
SparseLoCo & 3.12\% & \textbf{2.91} \\
DiLoCo & 100\% & 2.97 \\
DiLoCo-LOM+\textsc{Top-}$k$+EF & 3.12\% & 3.15 \\
DiLoCo+\textsc{Top-}$k$+EF & 3.12\% & 3.30 \\
\bottomrule
\end{tabular}
\end{table}

\section{Hyperparameter Selection Details}\label{app:sweeps}

Table~\ref{tab:opt-hp-registry} reports the hyperparameter search spaces for the 512M model size. We tune all methods at communication interval $H{=}15$ and reuse the best configurations for other settings; General and model architecture settings are fixed across all runs unless stated otherwise. In Table~\ref{tab:model-hparams-178M2B}, we provide architectural details for 178M and 2B model sizes. For 178M, we reduce the batch size to 32 leading to an effective batch size of $524,288$, and repeating the hyper-parameter sweeps as Table~\ref{tab:opt-hp-registry}, we observe the same optimal settings. For the 2B model size, we increase warmup to $800$ and perform a small sweep of learning-rates lower than the optimal setting of 178M and 500M models. Specifically, for DiLoCo we search \(\alpha_{\text{inner}}\!\in\!\{8\mathrm{e}{-}4,\,6\mathrm{e}{-}4\}\) and \(\alpha_{\text{outer}}\!\in\!\{0.6,\,0.4\}\), finding \(\alpha_{\text{inner}}{=}8\mathrm{e}{-}4\), \(\alpha_{\text{outer}}{=}0.6\) optimal; for SparseLoCo we search \(\alpha_{\text{inner}}\!\in\!\{1\mathrm{e}{-}3,\,8\mathrm{e}{-}4\}\) and \(\alpha_{\text{outer}}\!\in\!\{0.8,\,0.6\}\), finding \(\alpha_{\text{inner}}{=}1\mathrm{e}{-}3\), \(\alpha_{\text{outer}}{=}0.8\) optimal. For number of workers $R>8$ experiments, we ensure the same effective batch size used for $R{=}8$ by scaling the batch size accordingly. For all training runs we use Chinchilla optimal~\cite{hoffmann2022training} token budgets unless told otherwise.

\begin{table}[H]
\centering
\caption{Hyperparameter search spaces for the 512M-parameter model scale. Bold entries indicate the best settings. Model and general settings (top) are fixed across all runs. We tune all methods at $H{=}15$ and reuse the best hyperparameters when varying $H$. With higher number of workers $R$, DiLoCo's optimal setting remained the same whereas SparseLoCo enjoys slightly lower outer learning rate. The effective batch size is given per inner step across all workers.}
\label{tab:opt-hp-registry}

    \begin{minipage}[t]{0.49\textwidth}
        \centering
        \label{tab:general-hparams}
        \begin{tabular}{lc}
        \toprule
        \textbf{General Settings} & \textbf{Value} \\
        \midrule
        Token Budget & 10.26B \\
        Effective batch size & 4,194,304 \\
        Sequence length & 2048     \\
        Local batch size      & 256      \\
        Workers $R$ & 8     \\
        Warmup steps    & 500      \\
        Inner gradient clipping & 1.0 \\
        LR Decay & Cosine \\
        Inner optimizer & AdamW \\
        \bottomrule
        \end{tabular}
    \end{minipage}
    \hfill 
    \begin{minipage}[t]{0.49\textwidth}
        \centering
        \label{tab:model-hparams}
        \begin{tabular}{ll}
            \toprule
            \textbf{Parameter} & \textbf{Value} \\
            \midrule
            Total Parameters     & 512,398,848      \\
            Number of Layers     & 12        \\
            Hidden Size          & 1536       \\
            Intermediate Size & 5,440       \\
            Attention Heads      & 12        \\
            Vocabulary Size      & 32,000             \\ 
            FFN Activation       & SwiGLU              \\ 
            \bottomrule
        \end{tabular}
    \end{minipage}

\vspace{1em}

\begin{tabular}{lll}
\toprule
\textbf{Setting} & \textbf{Hyperparameter} & \textbf{Search Space} \\
\midrule
\multirow{2}{*}{AdamW Baseline}
& \(\alpha\) & 4e-4, 6e-4, 8e-4, 1e-3  \\
& & \textbf{2e-3}, \textbf{3e-3}, \textbf{4e-3}, 6e-3 \\
\midrule

\multirow{5}{*}{DiLoCo - Nesterov Outer} & \multicolumn{2}{l}{\textit{H=15}, $R\in\{8,16,32\}$} \\
\cmidrule(l){2-3}
& \(\alpha_{\text{inner}}\) & 4e-4, 6e-4, \textbf{8e-4}, 1e-3  \\
& \(\alpha_{\text{outer}}\) & 0.2, 0.4, \textbf{0.6}, 0.8, 1.0  \\
& momentum            & \textbf{0.9} \\
\midrule

\multirow{9}{*}{SparseLoCo (Density=0.78\%)} & \multicolumn{2}{l}{\textit{H=15}, $R{=}8$} \\
\cmidrule(l){2-3}
& \(\alpha_{\text{inner}}\) & 6e-4, 8e-4, \textbf{1e-3}, \textbf{2e-3}, 3e-3 \\
& \(\alpha_{\text{outer}}\) & 0.4, 0.6, 0.8, \textbf{1.0} \\
& error momentum (\(\beta\))       & 0.9, \textbf{0.95}, 0.999 \\
\cmidrule(l){2-3}
& \multicolumn{2}{l}{\textit{H=15}, $R{=}16$} \\
\cmidrule(l){2-3}
& \(\alpha_{\text{outer}}\) & 0.6, \textbf{0.8}, 1.0 \\
\cmidrule(l){2-3}
& \multicolumn{2}{l}{\textit{H=15}, $R{=}32$} \\
\cmidrule(l){2-3}
& \(\alpha_{\text{outer}}\) & 0.4, \textbf{0.6}, 0.8 \\

\midrule

\multirow{4}{*}{DiLoCo - SGD Outer}
& \(\alpha_{\text{inner}}\) & 6e-4, \textbf{1e-3} \\
& \(\alpha_{\text{outer}}\) & 0.8, \textbf{1.0} \\
& momentum            & \textbf{0.0} \\
\midrule

\multirow{4}{*}{DiLoCo-LOM}
& \(\alpha_{\text{inner}}\) & 6e-4, \textbf{8e-4}, 1e-3 \\
& \(\alpha_{\text{outer}}\) & 0.4, \textbf{0.6}, 0.8, 1.0 \\
& momentum            & \textbf{0.9} \\

\midrule
\multirow{2}{*}{DeMo}
& \(\alpha\) & 8e-4, \textbf{1e-3}, 3e-3 \\
& error momentum (\(\beta\))            & 0.95, \textbf{0.999} \\

\bottomrule
\end{tabular}
\end{table}

\end{document}